%% file: acl_latex.tex
\documentclass[11pt]{article}

\usepackage[preprint]{acl}

\usepackage{times}
\usepackage{latexsym}

\usepackage[T1]{fontenc}

\usepackage[utf8]{inputenc}

\usepackage{microtype}

\usepackage{inconsolata}

\usepackage{graphicx}

\usepackage{algorithm}
\usepackage{algorithmic}
\usepackage{amsmath}

\usepackage{newfloat}
\usepackage{listings}
\DeclareCaptionStyle{ruled}{labelfont=normalfont,labelsep=colon,strut=off} 
\floatstyle{ruled}
\newfloat{listing}{tb}{lst}{}
\floatname{listing}{Listing}
\usepackage{amsfonts}
\usepackage{multirow}
\usepackage{booktabs}

\title{On-Policy Self-Distillation with Teacher Evolution}

\author{%
Shufan Shen\textsuperscript{\rm 1,2}\quad
Zhongni Hou\textsuperscript{\rm 3,}\footnotemark[2]\quad
Chao Bi\textsuperscript{\rm 1}\quad 
Yufei Zhang\textsuperscript{\rm 3}\quad \\
\textbf{Guojun Yin\textsuperscript{\rm 3}\quad
Shuhui Wang\textsuperscript{\rm 1,}\footnotemark[2]}\\
$^{1}$ State Key Lab. of AI Safety, Institute of Computing Technology, \\ Chinese Academy of Sciences \\ $^{2}$ University of Chinese Academy of Sciences $^{3}$ Meituan \\
\texttt{shenshufan22z@ict.ac.cn}  }

\begin{document}
\maketitle
\renewcommand{\thefootnote}{\fnsymbol{footnote}}
\footnotetext[2]{Corresponding authors.}
\renewcommand{\thefootnote}{\arabic{footnote}}
\input{Sections/0_abstract}
\input{Sections/1_introduction}

\input{Sections/2_relatedwork}
\input{Sections/3_method}
\input{Sections/4_experiment}

\input{Sections/5_conclusion}

\bibliography{custom}

\clearpage

\appendix

\input{Sections/X_supplement}


\end{document}

%% file: Sections/0_abstract.tex
\begin{abstract}
On-Policy Self-Distillation (OPSD), which aligns a student model's token-level distribution with that of its self-teacher prompted by privileged context, has emerged as a promising direction to mitigate reward sparsity in sequence-level post-training paradigms. However, OPSD often fails to achieve continuous performance gains during training. In this paper, we find that the model's limited self-teaching capability underlies this phenomenon by generating unreliable guidance on certain samples. 
To address this problem, we propose Teacher-Evolution Self-Distillation~(TESD), which enables the self-teacher to generate rollouts and receive environmental feedback based on task correctness and context independence. First, using these rewards as training objectives, we perform teacher-oriented optimization to continuously enhance the model's self-teaching capability during distillation. Second, by estimating teacher guidance reliability from these rewards, we apply adaptive student-oriented optimization to down-weight samples with unreliable guidance. Extensive experiments across multiple datasets and model scales show that TESD consistently outperforms representative token-level and sequence-level methods, highlighting the necessity of improving supervision quality in OPSD. 
\end{abstract}

%% file: Sections/1_introduction.tex
\section{Introduction}

\begin{figure}[t]
  \centering
  \includegraphics[width=0.95\linewidth]{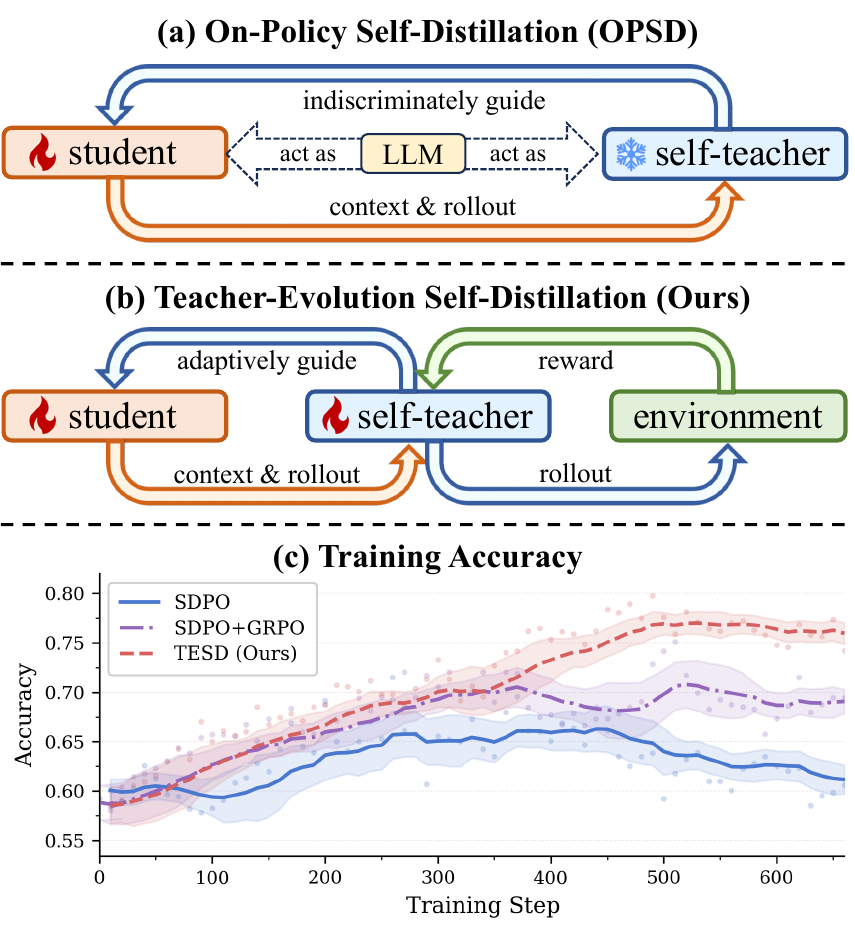}
  \vspace{-0.4cm}
  \caption{(a) In OPSD, the self-teacher provides guidance indiscriminately across all samples without improving its own capabilities, thereby degrading the quality of supervision delivered to the student.
  (b) TESD improves the self-teacher using the environmental rewards of its rollouts while delivering adaptive guidance by down-weighting samples with unreliable supervision.
  (c) TESD achieves more continuous performance improvement than current OPSD methods.
  }
  \label{fig:intro}
  \vspace{-0.4cm}
\end{figure}

Post-training has emerged as a crucial stage for enhancing the reasoning abilities of large language models (LLMs)~\cite{achiam2023gpt, kumar2025llm, zheng2025group, shen2026policy}. Among various post-training techniques, reinforcement learning with verifiable rewards~(RLVR) has proven to be a practical paradigm in domains where model outcomes can be automatically verified~\cite{shao2024deepseekmath}.
Despite achieving strong performance, representative RLVR methods assign the same reward to all tokens within a response. This sparse sequence-level scheme overlooks the varying significance of individual tokens, hindering the model’s ability to focus on important tokens during optimization~\cite{li2026rethinking}.

To address this limitation, On-Policy Self-Distillation~(OPSD) offers dense supervision by introducing a self-teacher as shown in Figure~\ref{fig:intro}~\cite{zhao2026reinforced, shenfeld2026self}. 
The self-teacher is constructed by prompting the student with a self-generated privileged context, which usually consists of a correct solution to the input query. 
By computing the per-token logits of the self-teacher on the student's rollouts, OPSD delivers token-level supervision while preserving the on-policy property. Therefore, it can maintain alignment between the dense supervision signal and the model’s current behavior~\cite{hubotter2026reinforcement}.

However, existing OPSD methods fail to continuously improve performance during training~\cite{yang2026self,li2026rethinking}, leading to suboptimal results as shown in Figure~\ref{fig:intro}(c). 
This phenomenon is correlated with their reliance on the assumption of the model's strong self-teaching ability, \textit{i.e.}, the self-teacher can always correctly guide diverse student rollouts based on the limited privileged context. 
\textit{First}, this assumption deviates from practical implementations. 
The self-teacher only has the additional privileged context compared to the student, without any improvements in model capability. Thus, the self-teacher may fail to associate the limited privileged context with diverse student rollouts~\cite{yu2026multi}, leading to unreliable guidance. This insufficient self-teaching capability also struggles to improve during training, as the current OPSD paradigm optimizes the model solely as a student.
\textit{Second}, relying on this unrealistic assumption, current methods overlook the teacher's potential to deliver unreliable guidance on certain samples. Consequently, the student is forced to follow the teacher's guidance across all samples indiscriminately~\cite{hubotter2026reinforcement}, thereby injecting misleading signals into its optimization process.
These two problems limit the quality of supervision received by the student, making it difficult for OPSD to achieve continuous performance improvement during the distillation process.

In this paper, we propose the Teacher-Evolution Self-Distillation~(TESD) framework to enhance the supervision quality of OPSD. As shown in Figure~\ref{fig:intro}(b), rather than generating rollouts only through the student, TESD enables the self-teacher to generate rollouts and receive rewards.
Based on these rewards, TESD introduces two complementary mechanisms to address the aforementioned problems: teacher-oriented optimization to improve the model's self-teaching ability, and adaptive student-oriented optimization to mitigate the impact of unreliable teacher guidance. 
Specifically, for the reward function, we design it with both task-correctness and context-independence to estimate the model's self-teaching ability. 
For teacher-oriented optimization, these rewards serve directly as training objectives for the self-teacher. This objective encourages the self-teacher to generate correct and distinct solutions from privileged contexts, thereby improving its ability to associate the privileged context with diverse student rollouts. 
For adaptive student-oriented optimization, these rewards reflect the reliability of teacher guidance across different samples and serve as sample-wise weights that adaptively scale the student's loss values. By suppressing gradient updates from samples with unreliable guidance, their negative impact on student optimization can be mitigated.

In experiments, we evaluate the effectiveness of TESD on pre-trained LLMs~\cite{yang2025qwen3, olmo2025olmo} with different parameter sizes and multiple downstream tasks~\cite{feng2024sciknoweval,tang2023toolalpaca, jain2025livecodebench}. The evaluation results show that TESD consistently outperforms current representative OPSD~\cite{hubotter2026reinforcement} and sequence-level RLVR~\cite{shao2024deepseekmath} methods. The superior performance of TESD demonstrates the significance of improving supervision quality during the self-distillation process. Our contributions are summarized as follows:

\begin{itemize}

    \item We introduce TESD, an OPSD framework that allows the self-teacher to generate rollouts and receive rewards regarding task correctness and context independence.

    \item With the rewards from teacher rollouts, we enhance OPSD's supervision quality through teacher-oriented and adaptive student-oriented optimization, explicitly improving the model across its dual roles as teacher and student.

    \item Extensive experiments demonstrate the superior performance of TESD, which reflects the importance of enhancing the supervision quality in the OPSD paradigm.
\end{itemize}

%% file: Sections/2_relatedwork.tex
\section{Related Work}

\subsection{RL with Verifiable Rewards}
RL with verifiable rewards~(RLVR) has been widely used to improve the reasoning ability of pre-trained LLMs~\cite{guo2025deepseek}. As a representative RLVR method, GRPO~\cite{shao2024deepseekmath} estimates group-relative advantages from sampled responses and removes the need for a separate critic model. Subsequent studies
further improve the GRPO by modifying the objective~\cite{chu2025gpg}, normalization~\cite{yu2025dapo}, and clipping~\cite{zhao2026reinforced}. 
Despite differences in training strategies and loss formulations, these methods uniformly utilize sequence-level rewards to optimize the model as a student, which is prompted solely by input queries. 
In contrast, we design a reward function incorporating task-correctness and context-independence terms to optimize the model as a self-teacher, which receives both the input query and privileged context. By optimizing the self-teacher with this reward, we leverage sequence-level feedback to enhance the supervision quality of the token-level OPSD.

\subsection{On-Policy Self-Distillation}
Unlike classical knowledge distillation, which forces the student to strictly fit the teacher's output trajectory~\cite{gu2024minillm}, on-policy distillation~(OPD) enables the teacher to provide guidance based on the student's trajectory~\cite{agarwal2024policy, lu2025onpolicydistillation}. This approach can transfer the teacher's token-level knowledge while preserving the distribution of the student, making it widely adopted in LLM post-training~\cite{yang2025qwen3,zeng2026glm}. Building upon this, the on-policy self-distillation~(OPSD) paradigm guides the student using a self-teacher, which is built by prompting the student with privileged contexts, thereby achieving OPD without the external teacher~\cite{hubotter2026reinforcement, zhao2026self, shenfeld2026self}. 
However, the OPSD paradigm fails to achieve performance improvement continuously during training. To tackle this problem, current studies suggest incorporating more privileged context~\cite{yu2026multi}, degrading OPSD as a weighted module of GRPO~\cite{yang2026self}, and implementing an entropy-based routing update strategy~\cite{jin2026entropy}. Different from these studies relying solely on student-generated rollouts, we empower the self-teacher to generate rollouts and obtain rewards from the environment. 
Using these rewards, we optimize the self-teacher and adaptively adjust the intensity of its sample-wise supervision, thereby achieving more consistent performance gains over existing methods through enhanced supervision quality.

%% file: Sections/3_method.tex
\section{Method}

\subsection{Preliminaries: On-Policy Self-Distillation}
Given an input query $x$, a privileged context $c$, and a student rollout $y^s$ sampled from the current policy $\pi_\theta(\cdot|x)$, the OPSD loss function is formulated as:
\begin{equation}
\begin{aligned}
    \mathcal{L}_{\text{S}}(\theta,x,y^s) = \frac{1}{|y^s|}\sum_{t=1}^{|y^s|} \mathbb{D} \Big( 
    \pi_{\theta}(\cdot \mid x, y^s_{<t})
    \\ \parallel \text{stopgrad}\big[\pi_\theta(\cdot \mid x, c, y^s_{<t})\big] \Big),
\end{aligned}
\label{eq:opsd}
\end{equation}
where $\text{stopgrad}[\cdot]$ denotes the stop-gradient operator and $|y^s|$ is the length of the rollout in tokens. $\mathbb{D}$ represents an arbitrary divergence metric~(\textit{e.g.}, KL-divergence or JS-divergence), which measures the distribution gap between the student and the self-teacher.
By computing a per-token distribution gap between the self-teacher and the student along the student’s own sampled rollouts, OPSD yields dense, token-level supervision in an on-policy manner.
In practice, the self-teacher is usually updated as an exponential moving average~(EMA) of the student during training~\cite{hubotter2026reinforcement}.

\subsection{Teaching Ability Investigation}

\begin{figure}[tb]
  \centering
  \includegraphics[width=0.95\linewidth]{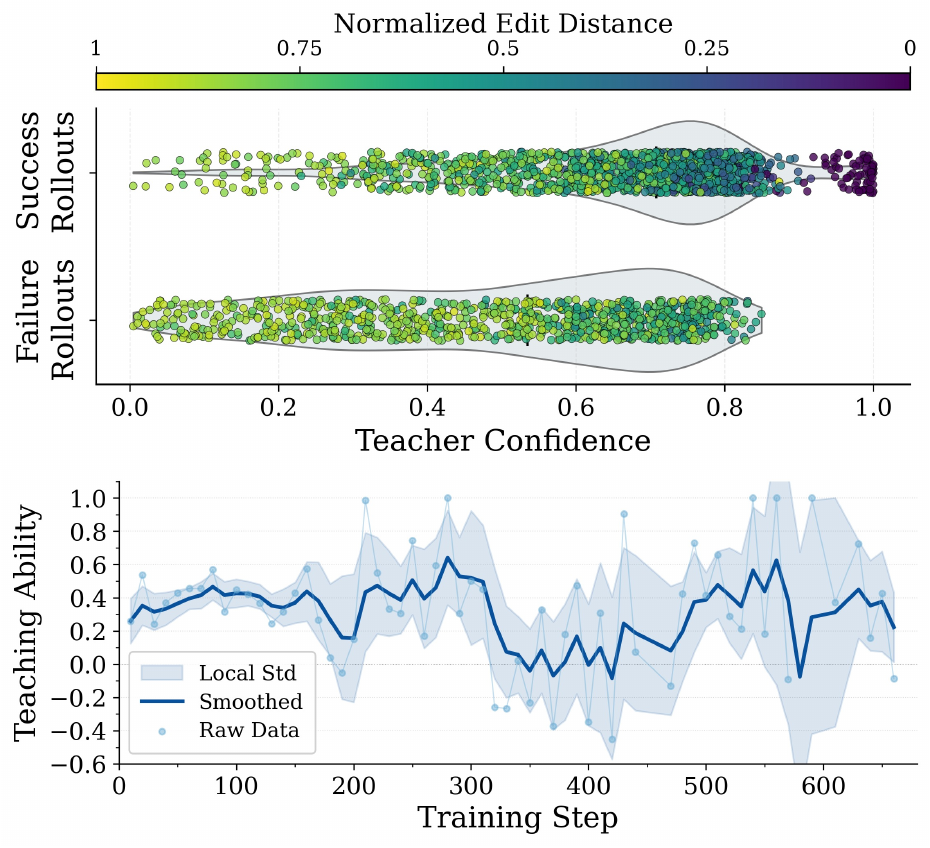}
  \vspace{-0.3cm}
  \caption{\textit{Top:} We plot the task correctness and teacher confidence (measured by mean token probability) of each student rollout, with the color representing its edit distance to the privileged context.
  \textit{Bottom:} The evolution of the model's teaching ability during the training process, which is estimated by the mean spearman correlation between teacher confidence and actual rewards. 
  Experiments are conducted on the Physics subset of SciKnowEval using Qwen3-8B. 
  }
  \label{fig:pre_study}
  \vspace{-0.3cm}
\end{figure}

\begin{figure*}[tb]
  \centering
  \includegraphics[width=1.0\linewidth]{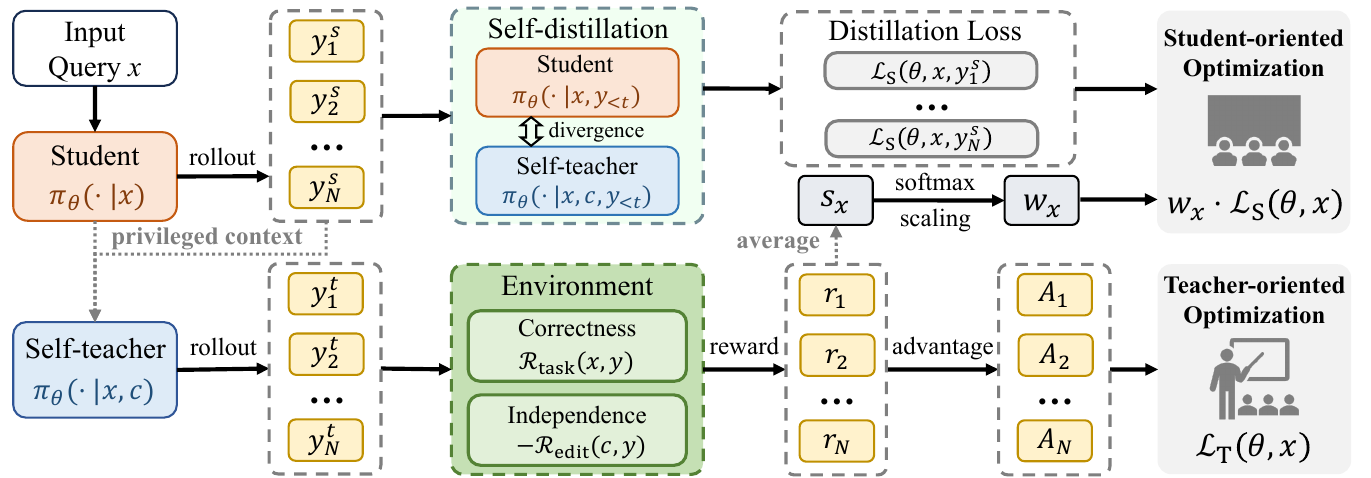}
  \vspace{-0.4cm}
  \caption{Overall framework of TESD. Given an input query, the student first generates rollouts, which provide the privileged context for the self-teacher. Then, TESD optimizes the student and the self-teacher, respectively. For the self-teacher, it interacts with the environment and learns from the obtained rewards of correctness and independence. For the student, it is supervised by the logits of the self-teacher. The loss value is adaptively weighted by the teacher's mean reward of the corresponding input query, which estimates the quality of teacher guidance.
  }
  \label{fig:framework}
  \vspace{-0.2cm}
\end{figure*}
Given that all supervision signals in the training objective of Equation~\ref{eq:opsd} are exclusively generated by the self-teacher $\pi_\theta(\cdot \mid x, c, y^s_{<t})$, its teaching ability is crucial to the success of the OPSD paradigm. 
At the top of Figure~\ref{fig:pre_study}, we investigate this ability via the correlation between the self-teacher's confidence in student rollouts and their verifiable rewards. This correlation reveals whether the self-teacher can assign high confidence to correct rollouts and low confidence to incorrect ones. Moreover, we visualize the similarity between each rollout and its privileged context. The normalized edit distance is selected as the similarity metric following previous studies~\cite{zheng2023judging}.
First, we find that the self-teacher struggles to distinguish the correctness of student rollouts, as it assigns low confidence~(0.0$\sim$0.5) to certain correct rollouts, while giving high confidence~(0.5$\sim$0.8) to a substantial number of incorrect ones.
Second, the self-teacher tends to assign higher confidence to rollouts that resemble the privileged context. This confidence-resemblance bias is supported by the observation that high-confidence rollouts tend to have smaller edit distances from the privileged context, while a large number of correct yet context-dissimilar rollouts are assigned low confidence.

Furthermore, we track the evolution of self-teaching ability during the training process at the bottom of Figure~\ref{fig:pre_study}. Rather than improving as the training progresses, the self-teacher's ability exhibits a downward trend coupled with significant instability. The results demonstrate that only student-oriented optimization in Equation~\ref{eq:opsd} is insufficient for improving the model's teaching ability. This stagnant or declining teaching ability acts as a bottleneck, preventing the OPSD from obtaining continuous performance improvements.

According to the above analysis, the self-teacher faces a challenge in providing reliable guidance on student rollouts, and this challenge does not diminish as training progresses. This phenomenon emphasizes the importance of improving the quality of supervision received by the student.

\subsection{Teacher-Oriented Optimization}
To improve the quality of guidance generated by the self-teacher, we enable it to interact with the environment. The rewards obtained from the environment are directly utilized to optimize the self-teacher, thereby continuously enhancing its teaching ability during the self-distillation process. Given a input query $x$ and a privileged context $c$, the self-teacher $\pi_{\theta}(\cdot|x,c)$ samples a set of $N$ teacher rollouts $\{y^t_1, y^t_2, \ldots, y^t_N\}$. Based on the insights into the confidence-resemblance bias shown at the top of Figure~\ref{fig:pre_study}, the reward function $r$ is defined as a combination of a verifiable task correctness and an edit-distance term:
\begin{equation}
    r_i=r(x, c, y^t_i) = \mathcal{R}_{\text{task}}(x, y^t_i) + \lambda_{\text{edit}} \mathcal{R}_{\text{edit}}(c, y^t_i),
\label{eq:reward}
\end{equation}
where $\mathcal{R}_{\text{task}} \in \{0, 1\}$ is the task correctness. $\mathcal{R}_{\text{edit}}\in[0,1]$ measures the normalized edit distance between the rollout and the privileged context, and $\lambda_{\text{edit}}$ is a balancing hyperparameter. 
As formulated in Equation~\ref{eq:reward}, the edit-distance reward encourages the self-teacher to generate rollouts distinct from the privileged context, thereby simulating scenarios where the teacher must generate responses conditioned on diverse student rollouts under a fixed context. Concurrently, the task-correctness reward encourages the teacher to provide token-level guidance that steers these trajectories toward correct solutions.

With this reward, we compute the group-wise advantage $A_i = \frac{r_i - \text{mean}(r)}{\text{std}(r)}$ following GRPO~\cite{shao2024deepseekmath}. The self-teacher is then optimized via the following objective:
\begin{equation}
    \mathcal{L}_{\text{T}}(\theta,x) = \frac{1}{N} \sum_{i=1}^{N} \min \left( \rho_i A_i, \text{clip}\left(\rho_i, 1\pm\epsilon\right) A_i \right),
\label{eq:teacher}
\end{equation}
where $\rho_i = \frac{\pi_\theta(y^t_i \mid x, c)}{\pi_{\text{old}}(y^t_i \mid x, c)}$ is the importance sampling ratio, $\pi_{\text{old}}$ represents the previous policy before the current training step. $\epsilon$ defines the clipping threshold to restrict policy updates.
This teacher-oriented optimization promotes rollouts that are correct and dissimilar to privileged contexts, while penalizing those that are incorrect or overly similar to privileged contexts. As a result, the self-teacher learns to assign high and low confidence to these two categories of student rollouts, respectively.
Moreover, continuously optimizing the self-teacher during the self-distillation process mitigates the degradation of teaching ability in the bottom of Figure~\ref{fig:pre_study}.

\subsection{Adaptive Student-Oriented Optimization}
With the proposed teacher-oriented optimization, we can gradually improve the ability of the self-teacher during the training process. However, it remains insufficient to alleviate the impact of unreliable teacher guidance on students during the early stages of training, when the self-teacher is still weak. 
To mitigate this problem, we propose an adaptive mechanism to identify queries with unreliable guidance and reduce their impact on the student-oriented optimization. 

Specifically, given an input query $x$ from the sampled batch $\mathcal{B}$, we compute the mean reward of the corresponding teacher rollouts to estimate the quality of teacher guidance:
\begin{equation}
    s_x = \frac{1}{N}\sum\limits_{i=1}^{N} r(x, c, y^t_i).
\label{eq:mean_reward}
\end{equation}
Based on this score, we can adjust the impact of different input queries on the model's update by weighting their loss values. Unlike hard-dropping methods that lead to insufficient training~\cite{yuan2023scaling}, we utilize a softmax normalization over the batch to compute the weights $w_x$. The normalized weights are scaled by the batch size $\vert \mathcal{B}\vert$ to preserve the overall scale of loss values.
\begin{equation}
    w_x = |\mathcal{B}| \cdot \frac{\exp(s_x)}{\sum_{x' \in \mathcal{B}} \exp(s_{x'})},
\label{eq:weight}
\end{equation}
During optimization, the distillation losses of different queries are adaptively weighted:
\begin{equation}
 \mathcal{L}_{\text{S}}(\theta, x) = w_x\cdot\frac{1}{N} \cdot \sum\limits_{i=1}^N \mathcal{L}_{\text{S}}(\theta,x,y^s_i),
\label{eq:student}
\end{equation}
where $\mathcal{L}_{\text{S}}(\theta,x,y^s_i)$ denotes the on-policy self-distillation loss in Equation~\ref{eq:opsd}, N represents the number of student rollouts.
Overall, the training objective of TESD on batch $\mathcal{B}$ is formulated as:
\begin{equation}
\mathcal{L}(\theta) = \frac{1}{|\mathcal{B}|}\sum_{x \in \mathcal{B}}\Big(\mathcal{L}_{\text{S}}(\theta,x)+\mathcal{L}_{\text{T}}(\theta, x)\Big)
\end{equation}
By integrating the teacher-oriented~(Equation~\ref{eq:teacher}) and adaptive student-oriented optimization~(Equation~\ref{eq:student}), TESD improves the self-teacher while mitigating the impact of unreliable guidance on the student. This joint formulation encourages the OPSD to achieve continuous performance gains.

%% file: Sections/4_experiment.tex
\section{Experiments}
\input{Tables/main_results}
\subsection{Experimental Setup}
\noindent\textbf{Models and Datasets}.
Qwen3-1.7B/4B/8B~\cite{yang2025qwen3} and Olmo3-7B-Instruct~\cite{olmo2025olmo} are selected as the base models. For datasets, we train models separately on six datasets following the prior work~\cite{hubotter2026reinforcement}. The first four datasets are question-answering benchmarks from the reasoning subset~(L3) of SciKnowEval~\cite{feng2024sciknoweval}, which cover scientific reasoning in Chemistry, Physics, Biology, and Materials Science. Moreover, we use a tool-use dataset from ToolAlpaca~\cite{tang2023toolalpaca}, where the model is required to map a tool-API specification and a user request to the correct tool call. Additionally, we adopt the v6 subset of LiveCodeBench~\cite{jain2025livecodebench}, a set of coding problems ranging from simple to competition-level. Each dataset is split into a training set and a test set.
Furthermore, we evaluate the model across several holdout tasks to measure the ability of TESD to mitigate catastrophic forgetting~\cite{Zhou2023InstructionFollowingEF,Verschueren2018Preprint,Lin2021TruthfulQAMH,Cobbe2021TrainingVT,Hendrycks2021MeasuringMP,Clark2018ThinkYH,Chen2021EvaluatingLL,Li2023HaluEvalAL}. 

\noindent\textbf{Baselines}.
We compare TESD with two baselines using the standard RLVR setting~\cite{hubotter2026reinforcement}. First, we adopt a strengthened implementation of GRPO in accordance with recent best practices~\cite{khatri2025art}. This implementation integrates recent refinements to GRPO, such as asymmetric clipping~\cite{yu2026dapo}, unbiased normalization~\cite{liu2025understanding}, and off-policy correction for efficient inference-based rollout generation~\cite{yao2025your}. Second, we utilize SDPO~\cite{hubotter2026reinforcement}, a representative OPSD method that functions in the same verifier-only RLVR setting as our method. SDPO provides the self-teacher with successful rollouts sampled by the current policy as the privileged context to offer token-level guidance. We do not incorporate other OPSD methods, like SDFT~\cite{shenfeld2026self}, due to their reliance on human-labeled solutions.
\input{Tables/lcbv6}

\noindent\textbf{Implementation Details}.
For fair comparisons, the number of generated rollouts is fixed at 16 across all methods. 
TESD allocates 8 rollouts each to the teacher and the student, whereas other methods generate all 16 rollouts using the student.
For evaluation, we use 16 sampled responses per problem and report the mean accuracy~(mean@16). Additional training and evaluation details are provided in the supplementary materials.

\input{Tables/ood}
\input{Tables/abl_teacher}

\subsection{Performance on Downstream Tasks}
\noindent\textbf{Learning without rich feedback}.
In Table~\ref{tab:main}, we evaluate the effectiveness of TESD on tasks without rich feedback, \textit{i.e.}, only one correct student rollout is selected as the privileged context. 
First, we observe that although SDPO is claimed to achieve faster convergence through the token-level supervision~\cite{hubotter2026reinforcement}, a substantial performance gap remains compared to the sequence-level GRPO upon full convergence~(\textit{e.g.}, 69.66\% vs. 73.40\% on Qwen3-4B). 
Second, when SDPO and GRPO are directly combined, the resulting performance lies between that of the two individual methods~(\textit{e.g.}, 72.27\% vs. 74.94\% on Qwen3-8B). This phenomenon indicates that a naive summation at the loss level fails to promote mutual enhancement. 
Finally, TESD outperforms all aforementioned baselines across different base models. 

\noindent\textbf{Learning with rich feedback}.
In Table~\ref{tab:lcbv6}, we present the performance of TESD and other baselines on the task with rich feedback, \textit{i.e.}, the privileged context including a correct student rollout and other code execution results such as runtime errors and failed unit tests. 
First, SDPO can outperform GRPO with rich feedback, reflecting the importance of the information density within the privileged context for OPSD. 
Second, combining SDPO and GRPO achieves the best performance on Qwen3-1.7B, yet lags behind TESD on larger model scales (Qwen3-4B/8B). This discrepancy stems from the limited capability of Qwen3-1.7B, where direct supervision of the student effectively accelerates its convergence. In contrast, for stronger models, integrating the self-teacher into the optimization framework improves the quality of supervision signals and then achieves better performance.
Finally, TESD consistently outperforms SDPO and GRPO across different base models, indicating that the effectiveness of TESD is compatible with the expansion of privileged context.

\input{Tables/abl_adaptive}

\noindent\textbf{Performance on Holdout Tasks}.
We evaluate the catastrophic forgetting of the model on other tasks after being trained with different methods on a specific task. As Table~\ref{tab:ood} shows, TESD learns the new task while mitigating the degradation of initial capabilities, achieving a better performance-forgetting tradeoff than SDPO and GRPO.

\subsection{Ablation Studies}
\noindent\textbf{Teacher-Oriented Optimization}.
We provide ablations of the teacher-oriented optimization on three datasets of the SciKnowEval benchmark using Qwen3-4B in Table~\ref{tab:abl_teacher}. First, applying sequence-level optimization to the student can improve the performance of OPSD~(67.06\% vs. 70.61\%), highlighting the necessity of leveraging environment feedback. 
Second, optimizing the self-teacher using sequence loss achieves performance improvements over OPSD~(67.06\% vs. 69.35\%), yet fails to outperform optimizing the student~(70.61\% vs. 69.35\%). This is because training only with a correctness reward encourages the teacher to copy the privileged context as the answer, thereby undermining its ability to guide diverse student rollouts.
By introducing a reward term based on edit distance in Equation~\ref{eq:reward}, this copying behavior can be effectively mitigated, thereby enhancing the teacher's ability and resulting in enhanced performance.

\begin{figure*}[tb]
  \centering
  \includegraphics[width=0.97\linewidth]{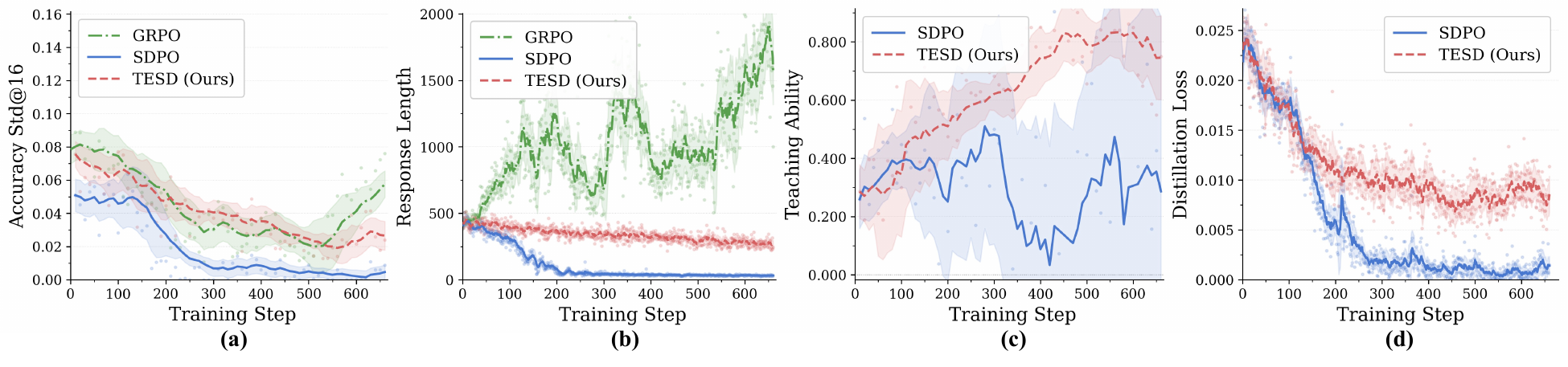}
  \vspace{-0.3cm}
  \caption{We report the evolution of the standard deviation of accuracy across 16 evaluations~(a), the model's average response length~(b), teaching ability~(c), and the distillation loss~(d) across the training process.
  }
  \label{fig:analysis_long}
  \vspace{-0.3cm}
\end{figure*}

\begin{figure}[tb]
  \centering
  \includegraphics[width=0.97\linewidth]{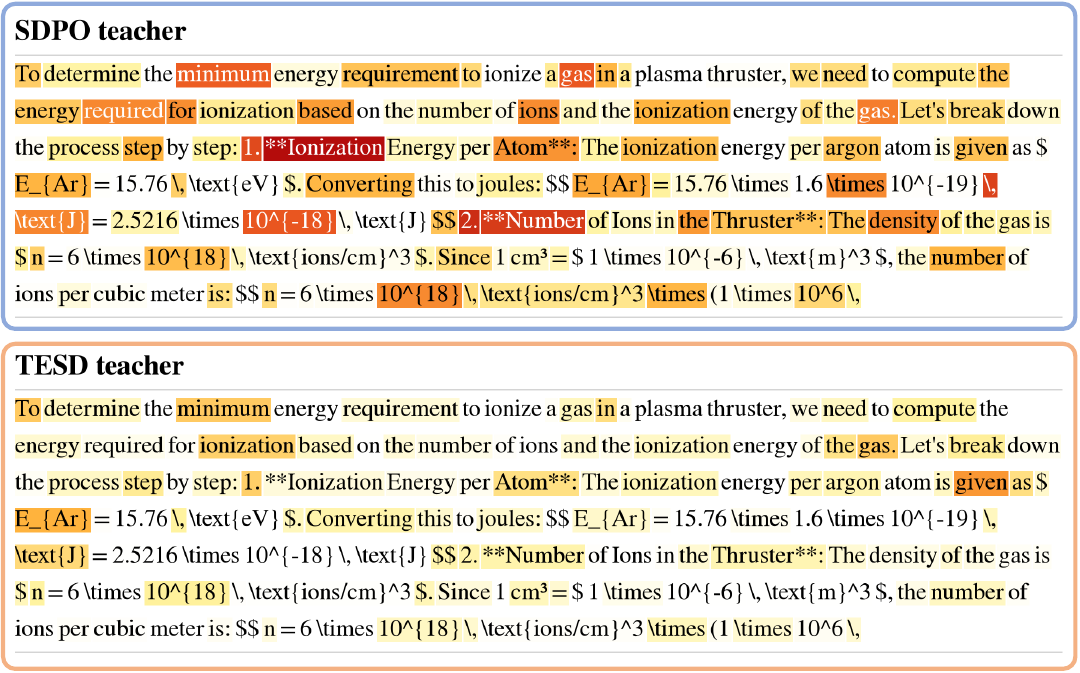}
  \caption{Token-level teacher credit heatmaps of SDPO and TESD with Qwen3-8B on the Physics dataset.}
  \label{fig:teacher_heatmap}
  \vspace{-0.5cm}
\end{figure}

\noindent\textbf{Adaptive Loss Weighting}.
As shown in Table~\ref{tab:abl_adaptive}, we investigate the effectiveness of the proposed weighting mechanism on three datasets of the SciKnowEval benchmark using Qwen3-8B. 
We observe that directly weighting the distillation loss of input queries using the mean reward in Equation~\ref{eq:reward} leads to a performance degradation~(69.53\% vs. 73.76\%). Introducing softmax normalization alleviates this issue, yet the performance remains worse than the unweighted baseline with teacher-oriented optimization~(70.56\% vs. 73.76\%). After incorporating scaling and clipping, a significant performance boost is achieved~(75.49\% vs. 70.56\%). This result not only validates the effectiveness of the proposed weighting mechanism but also reveals the importance of maintaining the overall scale of loss values during the weighting process.

Additional ablations of the rollout numbers and $\lambda_\textbf{edit}$ are provided in the supplementary materials.

\subsection{Analysis}

\noindent\textbf{Model Uncertainty}.
We explore the impact of TESD on the model's uncertainty through the accuracy standard deviation and response length of the model. Experiments are conducted on the Physics subset of SciKnowEval using Qwen3-8B.
First, as Figure~\ref{fig:analysis_long}(a) shows, both GRPO and TESD maintain a higher accuracy standard deviation compared to SDPO. 
Second, for response length in Figure~\ref{fig:analysis_long}(b), GRPO exhibits an increasing trend as training progresses, while SDPO shows a rapid decline in the early training stages before stabilizing. TESD falls between these two trajectories, presenting a gradual decrease in both metrics. 
These results indicate that, compared to SDPO, TESD tends to maintain more uncertainty in model outputs and prevents the over-confidence phenomenon during training. Meanwhile, TESD preserves a key characteristic of the OPSD paradigm: it attempts to generate shorter responses rather than relying on extended reasoning paths to boost performance like GRPO.

\noindent\textbf{Teaching Ability}.
For the teaching ability, we adopt the same metric as in Figure~\ref{fig:pre_study}. As shown in Figure~\ref{fig:analysis_long}(c), integrating the self-teacher into the optimization framework can gradually enhance its ability during training. Moreover, we find that the teaching ability in TESD exhibits greater stability~(as reflected by smaller local variance), while SDPO suffers from severe fluctuations.
Additionally, we present the distillation loss in Figure~\ref{fig:analysis_long}(d). 
Compared to SDPO, TESD shows higher losses during the middle and late stages of training. By optimizing the self-teacher, it can continuously provide significant guidance during training. 
As a result, the student is prevented from overfitting to an unoptimized self-teacher in the early stages, a pitfall that typically leads to performance stagnation in later phases.
Furthermore, we provide qualitative analysis in Figure~\ref{fig:teacher_heatmap}. In contrast to SDPO, which strongly supervises almost every token, TESD exhibits a sparser pattern. It diminishes the focus on common words like "a", "the", and "of", allowing the model to focus primarily on query-relevant keywords such as "ionization", "gas", and "E\_\{Ar\}".

\begin{figure}[tb]
  \centering
  \includegraphics[width=0.99\linewidth]{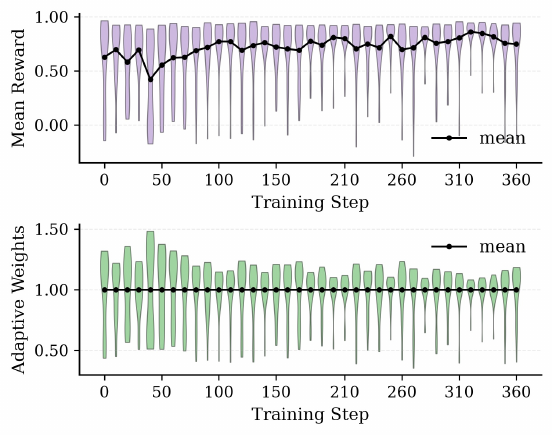}
  \vspace{-0.7cm}
  \caption{Distributions of the mean reward~(top) and the adaptive weight~(bottom) of the input batch.
  }
  \label{fig:analysis_weight}
  \vspace{-0.4cm}
\end{figure}

\noindent\textbf{Evolution of Adaptive Weights}.
In Figure~\ref{fig:analysis_weight}, we investigate the distribution of the mean teacher reward~(Equation~\ref{eq:mean_reward}) and the corresponding adaptive weight~(Equation~\ref{eq:weight}) within the input batch during training. 
In the early stages of training, due to the weak ability of the self-teacher, the adaptive mechanism decreases the loss of samples with unreliable teacher guidance to around 0.5. As teacher-oriented optimization improves the self-teacher over time, the teacher reward rises with the adaptive weights shifting toward 1.0. These results demonstrate the complementary relationship between the teacher-oriented and the adaptive student-oriented optimization throughout the training process.

Additional case studies and training time analysis are provided in the supplementary materials.

%% file: Tables/main_results.tex
\begin{table*}[t]
\centering
\small
\setlength\tabcolsep{12pt}
\begin{tabular}{l|l|ccccc|c}
\toprule
Model                     & Method    & Chemistry      & Physics        & Biology        & Material       & Tooluse        & Mean           \\ \midrule
\multirow{4}{*}{Qwen3-4B} & SDPO      & 80.92          & 71.48          & 56.63          & 73.07          & 66.18          & 69.66          \\
                          & GRPO      & \textbf{81.52} & \textbf{78.83} & 60.87          & 82.11          & 63.69          & 73.40          \\
                          & SDPO+GRPO & 79.58          & 75.37          & 54.62          & 81.85          & 62.87          & 70.86          \\
                          & TESD~(Ours)      & 80.30          & 77.50          & \textbf{63.50} & \textbf{83.31} & \textbf{67.00} & \textbf{74.32} \\ \midrule
\multirow{4}{*}{Qwen3-8B} & SDPO      & 77.83          & 67.73          & 62.63          & 79.99          & 62.87          & 70.21          \\
                          & GRPO      & 80.71          & 80.47          & 63.13          & \textbf{82.11} & 68.29          & 74.94          \\
                          & SDPO+GRPO & 77.62          & 76.80          & 60.25          & 79.32          & 67.37          & 72.27          \\
                          & TESD~(Ours)      & \textbf{81.55} & \textbf{80.55} & \textbf{64.13} & 81.78          & \textbf{70.02} & \textbf{75.61} \\ \midrule
\multirow{4}{*}{Olmo3-7B} & SDPO      & 79.73               & 56.25               &  \textbf{61.37}              & 73.27               &  63.60              &  66.84              \\
                          & GRPO      &  78.42              &  67.27              & 55.37               & 76.99               &  \textbf{65.60}              &  68.42              \\
                          & SDPO+GRPO &   78.69             &  66.02              &  52.50              &  75.47              &  62.32              & 67.00               \\
                          & TESD~(Ours)      &  \textbf{80.30}              &  \textbf{73.98}              & 54.50               &  \textbf{77.39}              & 62.59               & \textbf{69.75}               \\ \bottomrule
\end{tabular}
\caption{Performance comparisons on SciKnowEval and Tooluse benchmarks. All methods are trained to convergence, and scores are reported as the maximum mean@16 accuracy~(\%) achieved during training.}
\label{tab:main}
\vspace{-0.3cm}
\end{table*}

%% file: Tables/lcbv6.tex
\begin{table}[t]
\centering
\small
\setlength\tabcolsep{2pt}
\begin{tabular}{@{}l|ccc|c@{}}
\toprule
Method & Qwen3-1.7B     & Qwen3-4B       & Qwen3-8B       & Mean           \\ \midrule
SDPO   & 45.35          & 56.97          & 62.38          & 54.90          \\
GRPO   & 44.96          & 56.12          & 60.85          & 53.98          \\
SDPO+GRPO   & \textbf{46.38}          & 59.88          & 61.95          & 56.07          \\
TESD~(Ours)   & 45.98 & \textbf{60.07} & \textbf{63.31} & \textbf{56.45} \\ \bottomrule
\end{tabular}
\caption{Performance comparisons on LCBv6 benchmark using different base models.}
\label{tab:lcbv6}
\vspace{-0.3cm}
\end{table}

%% file: Tables/ood.tex
\begin{table*}[t]
\centering
\small
\setlength\tabcolsep{3pt}
\begin{tabular}{@{}l|c|ccccccccc@{}}
\toprule
       & Task           & \multicolumn{9}{c}{Holdout Tasks}                                                                                                                    \\ \midrule
Method & Physics          & IFEval         & MMLU-Pro       & TruthfulQA     & GSM8K          & Math500 & ARC-Challenge  & HumanEval      & \multicolumn{1}{c|}{HaluEval} & Mean \\ \midrule
Base   & 58.44          & 87.05          & 73.70          & 86.05          & 91.89          & 82.13       & 90.53          & 64.02          &  \multicolumn{1}{c|}{75.93}         & 81.41    \\ \midrule
SDPO   & 65.08          & 86.33          & 68.70          & 87.64 & 87.19          & 74.60       & 89.33          & 65.85 & \multicolumn{1}{c|}{75.68}    & 79.42    \\
GRPO   & 78.75          & 87.65          & 71.70 & 85.68          & 91.28          & 79.20       & 89.16          & 64.63          & \multicolumn{1}{c|}{76.01}    & 80.66    \\
TESD  & \textbf{79.24} & 86.21 & 72.70 & 88.98          & 91.51          & 80.40 & 90.36        & 67.68          & \multicolumn{1}{c|}{76.31}    & \textbf{81.77}    \\ \bottomrule
\end{tabular}
\vspace{-0.1cm}
\caption{Performance on avoiding catastrophic forgetting. We compare the base model~(Qwen3-8B) and the final checkpoint of GRPO, SDPO and TESD on the training task~(Physics) and holdout tasks.}
\label{tab:ood}
\vspace{-0.2cm}
\end{table*}

%% file: Tables/abl_teacher.tex
\begin{table}[t]
\centering
\small
\begin{tabular}{@{}l|ccc|c@{}}
\toprule
Method               & Physics & Biology & Material & Mean \\ \midrule
Baseline             &  71.48       & 56.63        & 73.07         & 67.06     \\
+Seq-level Loss &  75.37       &  54.62       & 81.85         & 70.61     \\
+Priv-Context  & 73.20        &  51.88       & 82.98         & 69.35     \\
+Edit-Reward  &  76.48       & 63.75        & 81.65         &  \textbf{73.96}    \\ \bottomrule
\end{tabular}
\caption{Ablations of the teacher-oriented optimization. Based on SDPO, we sequentially incorporating sequence-level loss~(+Seq-level Loss), optimizing the self-teacher with privileged context~(+Priv-Context), and adopting an edit-distance reward~(+Edit-Reward).
}
\label{tab:abl_teacher}
\end{table}

%% file: Tables/abl_adaptive.tex
\begin{table}[t]
\centering
\small
\begin{tabular}{@{}l|ccc|c@{}}
\toprule
Method            & Physics & Biology & Material & Mean \\ \midrule
Baseline          &  77.66       &  62.50       &  81.12        & 73.76     \\
+Mean Reward &  74.84       &  55.37       &  78.39        &  69.53    \\
+Softmax          &  73.67       & 58.75        & 79.26         & 70.56     \\
+Scaling    & 80.55        &  64.13       & 81.78         & \textbf{75.49}     \\ \bottomrule
\end{tabular}
\caption{Ablations of the adaptive weighting mechanism. We evaluate the impact of weighting the query losses based on the mean reward of teacher rollouts~(+Mean Reward), normalizing these rewards via softmax~(+Softmax), and scaling the normalized results~(+Scaling).}
\label{tab:abl_adaptive}
\vspace{-0.2cm}
\end{table}

%% file: Sections/5_conclusion.tex
\section{Conclusion}
In this paper, we propose a method called Teacher-Evolution Self-Distillation~(TESD) to achieve on-policy self-distillation with continuous performance improvements by enhancing the quality of the self-teacher's supervision. 
First, we design a reward function considering both correctness and independence to evaluate the self-teacher's ability to guide diverse student rollouts. Then, using this reward, we perform teacher-oriented optimization to improve the model's self-teaching capability during training. Finally, we utilize this reward for adaptive student-oriented optimization, which suppresses updates on queries with potentially unreliable teacher supervision.
Extensive experiments indicate the stronger performance of TESD compared to representative OPSD methods, indicating the importance of enhancing the supervision quality.
In future, we will investigate the optimization geometry of TESD and apply it to other domains such as multimodal reasoning~\cite{shen2026vl}.

%% file: Sections/X_supplement.tex
\section{Supplement Material}

\subsection{Implementation Details}
\begin{figure}[t]
  \centering
  \includegraphics[width=1.0\linewidth]{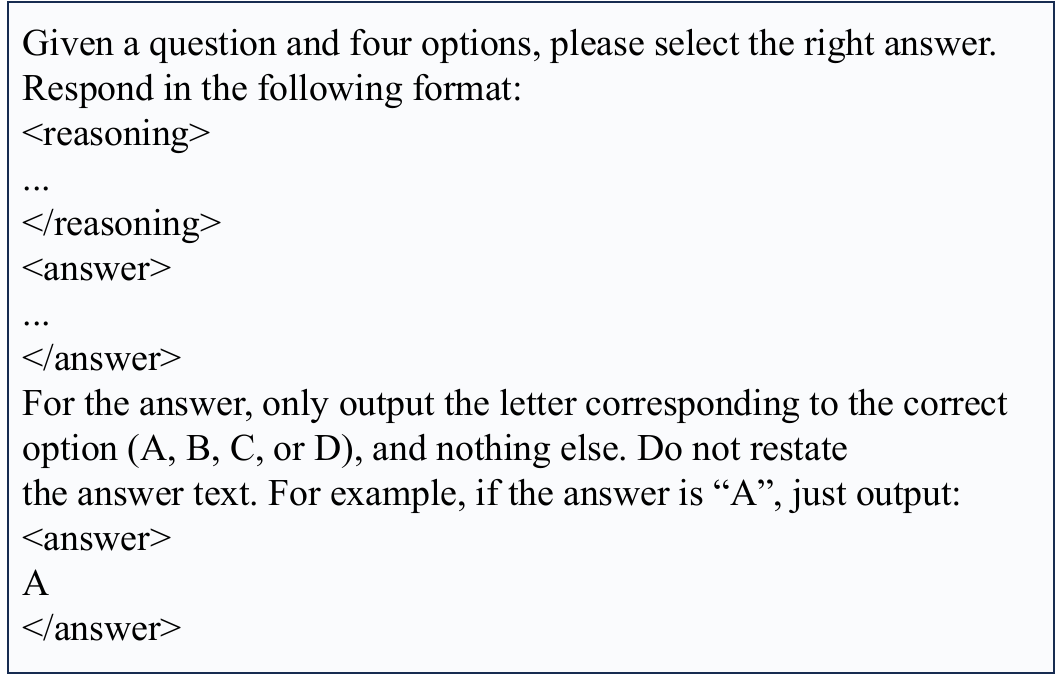}
  \caption{System prompt for ScienceQA.
  }
  \label{fig:qa_prompt}
\end{figure}

\begin{figure}[t]
  \centering
  \includegraphics[width=1.0\linewidth]{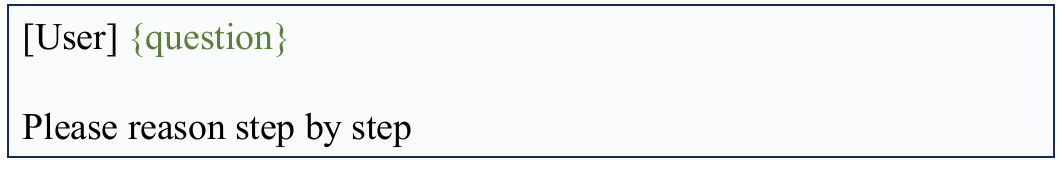}
  \caption{User prompt for ScienceQA.
  }
  \label{fig:qa_user}
\end{figure}

\begin{figure}[t]
  \centering
  \includegraphics[width=1.0\linewidth]{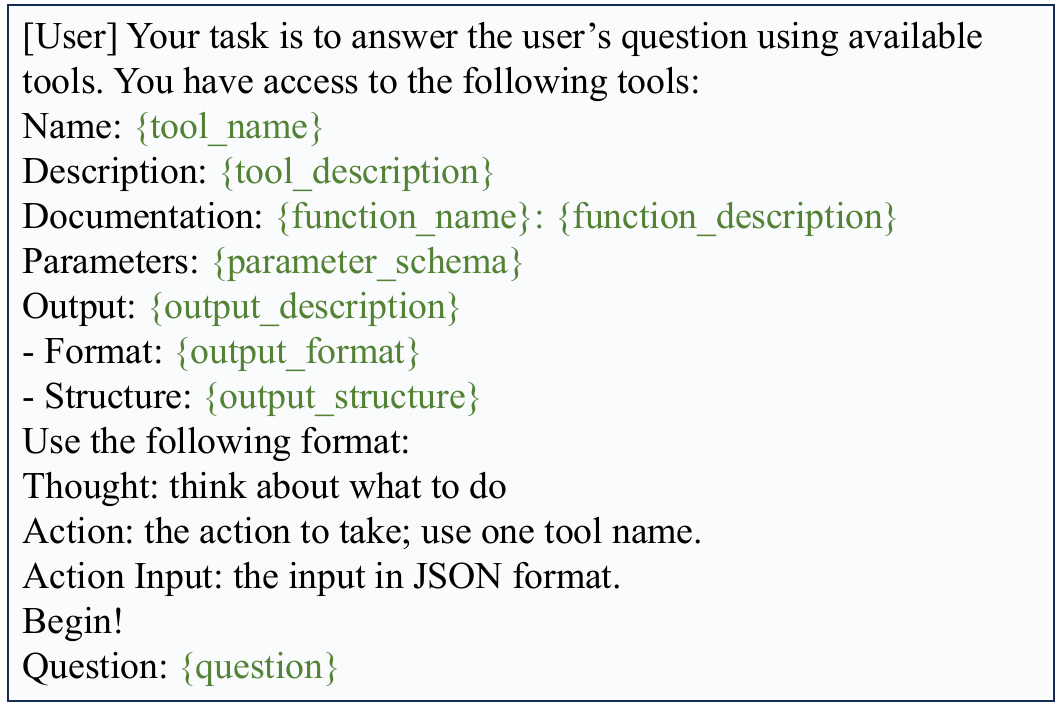}
  \caption{User prompt for ToolUse.
  }
  \label{fig:tool_user}
\end{figure}

\begin{figure}[t]
  \centering
  \includegraphics[width=1.0\linewidth]{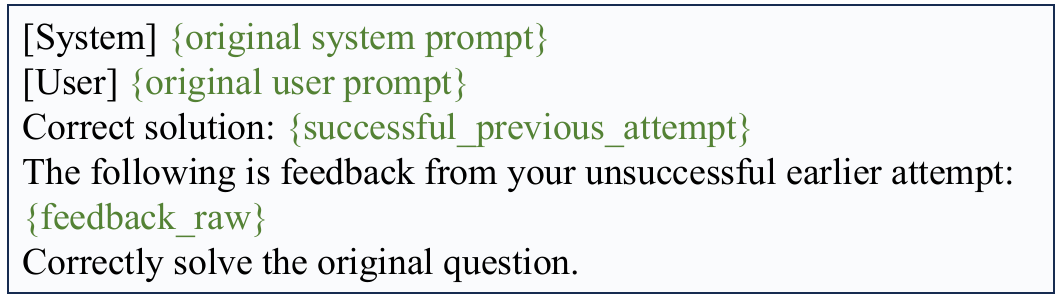}
  \caption{Teacher input template.
  }
  \label{fig:teacher_prompt}
\end{figure}

\subsubsection{Training Strategies}

All methods are implemented on top of the verl framework~\cite{sheng2024hybridflow} and trained with FSDP on a single 8$\times$H20 node. All training rollouts use temperature $T{=}1.0$, top-$p{=}1.0$, top-$k{=}{-}1$ with sampling enabled. The thinking mode of the Qwen3 family is disabled. Inputs and responses are bounded by 2048 and 8192 tokens, respectively. 
All methods use AdamW ($\beta_1{=}0.9,\ \beta_2{=}0.999$) with weight decay $0.01$, gradient clipping at $1.0$, a constant learning-rate schedule with $10$ warmup steps, and $1$ PPO epoch per iteration. The training batch size is $32$ prompts per step. Checkpoint and validation frequency are set to every $200$ and $10$ training steps, respectively. We train the model for 30 epochs on both the science question-answering tasks and for 15 epochs on the ToolUse task. For LCBv6, we train the model for 40 steps across all methods.

For GRPO, we sample $16$ rollouts per prompt with a mini-batch size of $8$. The policy-gradient loss adopts asymmetric PPO clipping with $\epsilon_{\text{low}}{=}0.2$, $\epsilon_{\text{high}}{=}0.28$ and a dual-clip lower bound $c{=}3.0$. To support efficient inference-based rollout generation, we apply token-level truncated importance-sampling (IS) correction with a threshold of $2.0$. No KL penalty is used.
The learning rate is selected from 1e-5 and 1e-6 based on validation performance.

For SDPO, we adopt its official implementations~\cite{hubotter2026reinforcement}. The self-teacher is maintained as an exponential moving average~(EMA) of the student with an update rate of $0.05$. The distillation loss is the generalized Jensen--Shannon divergence over the full logit distribution, approximated by the top-$100$ logits with an explicit tail bucket. The distillation importance-sampling ratio is clipped at $2.0$. The same asymmetric clipping and token-level IS correction as in the GRPO baseline are applied to the student policy. Successful student rollouts are selected as the privileged context, with the self-success exclusion enabled and thinking traces removed from the demonstration.

For TESD, it samples $8$ student rollouts and $8$ teacher rollouts. The teacher reward follows Eq.~(2): task correctness $R_{\text{task}}\in\{0,1\}$ minus an independence reward $\lambda_{\text{edit}}\,R_{\text{edit}}$. $R_{\text{edit}}\in[0,1]$ is the token-level normalized edit distance computed using the \texttt{SequenceMatcher} of difflib package in Python. $\lambda_{\text{edit}}$ is set as 1.0 across all experiments. The EMA teacher update rate is reduced to $0.01$ for stabilizing the training process. For the adaptive student-oriented optimization (Eq.~5--7), the weight $w_x$ is computed with temperature $\tau{=}1.0$, then clipped to $[0.05,5.0]$. Table~\ref{tab:hyper} summarizes the method-specific hyperparameters.

\begin{table*}[h]
\centering
\begin{tabular}{lccc}
\toprule
\textbf{Hyperparameter} & \textbf{GRPO} & \textbf{SDPO} & \textbf{TESD} \\
\midrule
Rollouts / prompt (total) & 16 & 16 & 16 \\
\ \ student rollouts & --- & 16 & 8 \\
\ \ teacher rollouts & --- & 0 & 8 \\
PPO mini-batch size & 8 & 32 & 32 \\
Advantage estimator & GRPO & GRPO & GRPO \\
Policy loss & student PG & KL-Div & KL-Div + teacher PG \\
Clip $\epsilon_{\text{low}}/\epsilon_{\text{high}}$ & \multicolumn{3}{c}{$0.2\,/\,0.28$} \\
Dual-clip $c$ & \multicolumn{3}{c}{$3.0$} \\
IS correction (token, thr.) & \multicolumn{3}{c}{token, $2.0$} \\
\midrule
Distillation $\alpha$ (JSD) & --- & 0.5 & 0.5 \\
Distillation top-$k$ & --- & 100 & 100 \\
Distillation IS clip & --- & 2.0 & 2.0 \\
Teacher regularization & --- & EMA & EMA \\
EMA update rate $\tau$ & --- & 0.05 & 0.01 \\
\midrule
Edit Reward $\lambda_{\text{edit}}$ & --- & 0 & 1.0 \\
Gate strategy & --- & --- & reward\_mean \\
Gate normalization & --- & --- & softmax \\
\bottomrule
\end{tabular}
\caption{Hyperparameter settings of different methods.}
\label{tab:hyper}
\end{table*}

\subsubsection{Evaluation Strategies}

For in-distribution evaluation, we evaluate the model on the test split of each task every $10$ training steps. 
For each test problem, we sample $16$ responses with a temperature of $0.6$ and top-$p = 0.95$, and report the mean accuracy (mean@16), \textit{i.e.}, the fraction of correct responses averaged over the $16$ samples. The final reported metric is the maximum mean@16 observed across the training trajectory, ensuring a fair comparison of convergence upper bounds. The task correctness reward $R_{\text{task}}$ is rule-based and task-specific:
\begin{itemize}
    \item \textbf{SciKnowEval}~(multiple-choice questions): The model is prompted to output the option letter inside the \texttt{<answer>...</answer>} tags via a system instruction. We extract the letter inside the tags and set $\mathcal{R}_{\text{task}} = 1$ if it matches the ground-truth option, and $0$ otherwise.

    \item \textbf{ToolUse}: Given a tool-API specification and a user request, the model generates a Thought-Action-Action\_Input trace. We set $R_{\text{task}}=1$ if the model prediction matches the ground-truth tool call, and $0$ otherwise.

    \item \textbf{LCBv6}~(coding problems): The model outputs a Python program in a fenced code block. We set $R_{\text{task}} = 1$ if the program passes all hidden test cases within the time limit in a sandbox environment, and $0$ otherwise.
\end{itemize}

For hold-out task evaluation in Table~3 of the main paper, we evaluate the final converged checkpoint of each method on eight hold-out tasks.
For IFEval, MMLU-Pro, TruthfulQA, ARC-Challenge, HumanEval, and HaluEval, we utilize the \texttt{lm-evaluation-harness}~\cite{eval-harness} framework with a generation budget of $32768$ tokens and greedy decoding.
We apply the model's default settings for the chat template, except for HumanEval, where generation is performed in the raw completion format. ARC-Challenge is evaluated in the multiple-choice setting. Regarding the evaluation metrics, IFEval reports strict instruction-following accuracy, MMLU-Pro and TruthfulQA report exact-match accuracy, HumanEval reports \texttt{pass@1}, and HaluEval reports exact-match accuracy for hallucination avoidance on the dialogue subset.
For mathematical reasoning tasks~(MATH-500 and GSM8K), we evaluate using a temperature of $0.6$ and top-$p = 1.0$, reporting \texttt{pass@1}. Answers are extracted and normalized before comparison with the ground truth.

\begin{table}[h]
\centering
\begin{tabular}{lcc}
\toprule
\textbf{Dataset} & \textbf{Train} & \textbf{Test} \\
\midrule
SciKnowEval-Biology    & 450  & 50  \\
SciKnowEval-Physics    & 720  & 80  \\
SciKnowEval-Chemistry  & 1890 & 210 \\
SciKnowEval-Materials  & 841  & 94  \\
ToolUse (ToolAlpaca)   & 4046 & 68  \\
LCBv6 (LiveCodeBench)  & 131  & 131 \\
\bottomrule
\end{tabular}
\caption{Train/test sizes of the six tasks.}
\label{tab:data}
\end{table}

\begin{figure*}[t]
  \centering
  \includegraphics[width=1.0\linewidth]{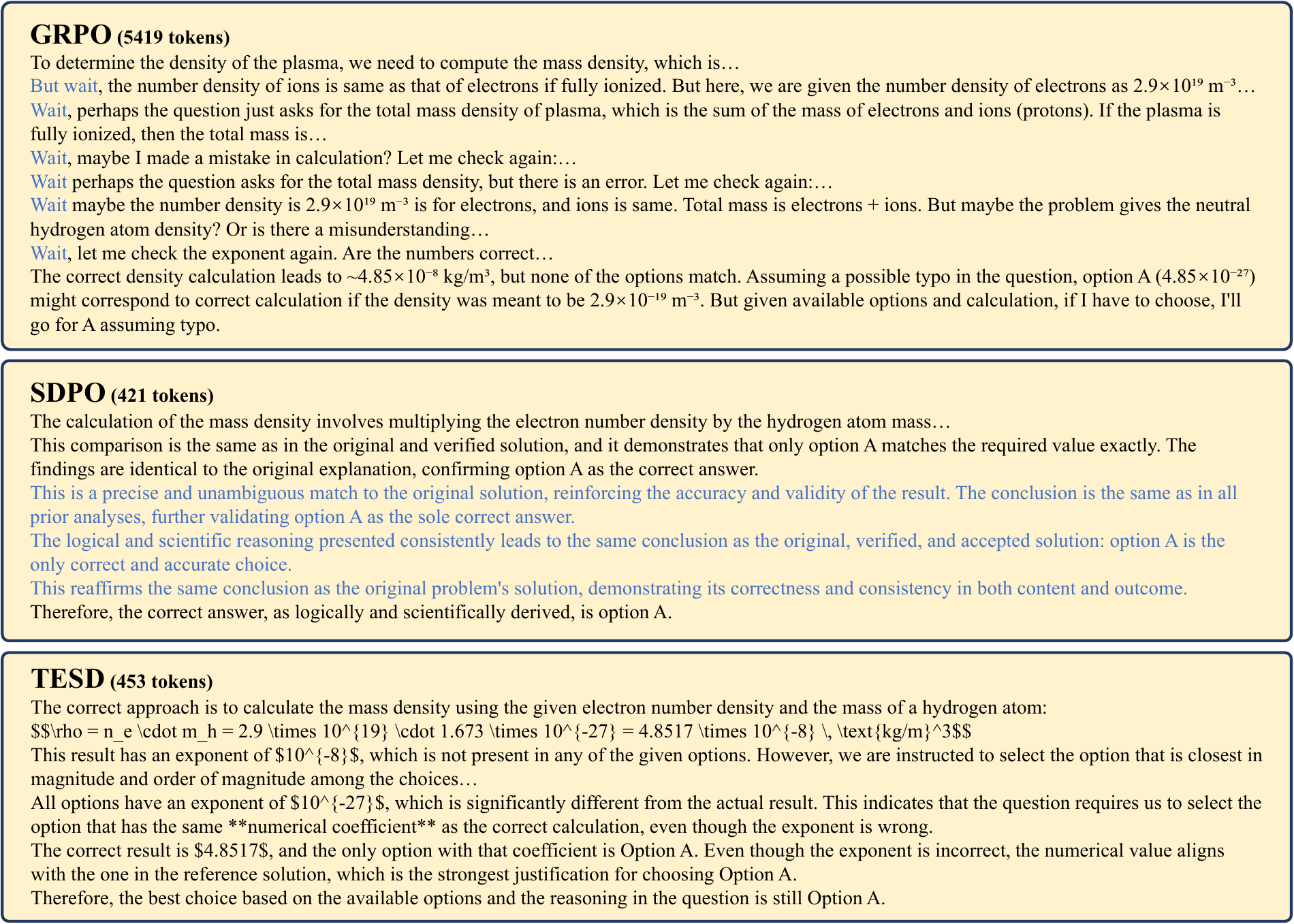}
  \caption{Case studies of different methods on the Physics subset of the SciKnowEval benchmark using Qwen3-8B.
  }
  \label{fig:cases}
\end{figure*}

\subsubsection{Prompt Templates}
Following SDPO~\cite{hubotter2026reinforcement}, we use the same dataset prompts for student rollout and evaluation. For Qwen models, we disable the thinking-mode chat-template option during prompt construction. The concrete ScienceQA and ToolUse prompt templates are shown in Figures~\ref{fig:qa_prompt},~\ref{fig:qa_user}, and~\ref{fig:tool_user}. The prompt used as input to the self-teacher is
shown in Figure~\ref{fig:teacher_prompt}.

\subsubsection{Normalized Edit Distance}
For each teacher rollout $y$ and its associated privileged-context solution $c$, we first decode the rollout while removing special tokens. We then lowercase both strings and split them on whitespace, obtaining token sequences $\tilde y$ and $\tilde c$. 
We compute their similarity using Python's \texttt{difflib.SequenceMatcher.ratio()} with its default settings. Let $M(\tilde y,\tilde c)$ be the total number of tokens in the matching blocks identified by \texttt{SequenceMatcher}. The normalized similarity and its complementary distance are
\[
\begin{aligned}
S_{\mathrm{edit}}(y,c)
&=\frac{2M(\tilde y,\tilde c)}{|\tilde y|+|\tilde c|},\\
D_{\mathrm{edit}}(y,c)
&=1-S_{\mathrm{edit}}(y,c).
\end{aligned}
\]
Thus, identical token sequences have $S_{\mathrm{edit}}=1$ and $D_{\mathrm{edit}}=0$, whereas sequences with no matching tokens have $S_{\mathrm{edit}}=0$ and $D_{\mathrm{edit}}=1$.

\subsection{Additional Experiments}

\subsubsection{Scaling of Rollouts}
\input{Tables/rollout}
In Table~\ref{tab:rollout}, we evaluate the efficiency of SDPO and TESD in leveraging scaling rollouts to achieve performance improvements. 
The performance of SDPO does not improve significantly with the increasing number of rollouts~(\textit{e.g.}, when $N$ increases from 32 to 48, the performance drops from 55.25\% to 55.12\%). In contrast, the performance of TESD improves consistently as the number of rollouts grows. This demonstrates that, by allocating partial rollout computation to the self-teacher, TESD can more effectively leverage increased training time to yield consistent performance gains.

\subsubsection{Qualitative Results}
In Figure~\ref{fig:cases}, we present the responses generated by models trained with different methods. 
First, compared to OPSD methods, GRPO generates significantly longer responses that contain a large amount of redundant reasoning (such as repetitive "wait" tokens), leading to wasted computations. Second, while SDPO produces shorter answers, it lacks genuine reasoning depth, dedicating numerous tokens to excessive self-affirmations. This phenomenon highlights the overconfidence issue inherent in existing OPSD paradigms~\cite{li2026rethinking}. Finally, in contrast to both methods, TESD successfully solves the problems with concise responses dominated almost entirely by effective, substance-driven reasoning.

\subsubsection{Effects of $\lambda_\text{edit}$}
In Table~\ref{tab:lambda_edit}, we investigate the impact of the hyperparameter $\lambda_\text{edit}$ in Equation~\ref{eq:reward} on the performance of TESD.
First, we observe that different subsets require different optimal values for $\lambda_{\text{edit}}$. For instance, the optimal $\lambda_{\text{edit}}$ is 1.0 for the biology subset and 0.7 for the materials subset. Second, setting $\lambda_{\text{edit}} = 1.0$ achieves the highest average performance across all three subsets. Therefore, to avoid dataset-specific hyperparameter tuning, we fix $\lambda_{\text{edit}} = 1.0$ for all main experiments.

\begin{table}[]
\centering
\begin{tabular}{@{}c|ccc|c@{}}
\toprule
$\lambda_\text{edit}$ & Biology & Material & Physics & Mean  \\ \midrule
0.1    & 51.88   & 78.99    & \textbf{76.48}   & 69.12 \\
0.3    & 53.75   & 79.45    & 76.25   & 69.82 \\
0.5    & 54.00   & 79.92    & 75.23   & 69.72 \\
0.7    & 50.88   & \textbf{82.78}    & 73.05   & 68.90 \\
1.0    & \textbf{54.87}   & 81.65    & 75.37   & \textbf{70.63} \\ \bottomrule
\end{tabular}
\caption{Ablation studies on the hyperparameter $\lambda_\text{edit}$. Experiments are conducted on SciKnowEval benchmark using Qwen3-4B over 10 training epochs.}
\label{tab:lambda_edit}
\end{table}

\subsubsection{Training Efficiency}
\begin{figure}[t]
  \centering
  \includegraphics[width=1.0\linewidth]{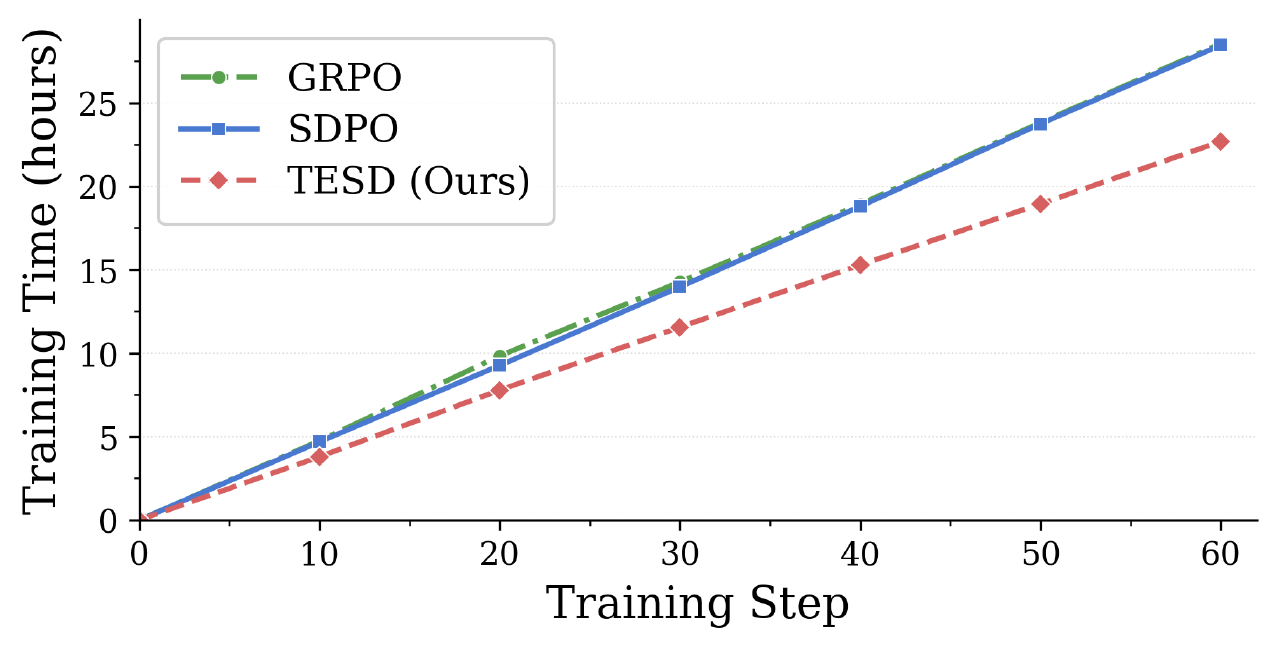}
  \caption{The time cost of different methods during the training process. Experiments are conducted on the LCBv6 benchmark using Qwen3-8B.
  }
  \label{fig:training_time}
\end{figure}

In Figure~\ref{fig:training_time}, we compare the time cost of different methods on the LCBv6 benchmark. 
We observe that as training progresses, TESD requires increasingly less time compared to GRPO and SDPO. This efficiency gain arises because a substantial portion of the training time in coding tasks is consumed by code execution and sandbox evaluation. Since TESD better enhances the model's capabilities throughout the training process, it generates programs that are more structured and efficient to execute, thereby accelerating training speed. 

\subsection{Discussion}

\subsubsection{Relation with Methods  improving Privileged Context}
Recently, several concurrent studies have improved OPSD from the perspective of privileged context by increasing its volume~\cite{yu2026multi} or strengthening it through contrastive exemplars~\cite{pan2026rlcsd}. In this section, we discuss the similarities and differences between them.
\begin{itemize}
    \item \textit{Similarity}. Both TESD and these concurrent methods identify the quality of teacher guidance as the primary bottleneck that constrains the performance of the OPSD paradigm, and actively seek to improve it.
    \item \textit{Difference}. While these concurrent methods focus heavily on the quality of the context provided to the self-teacher, TESD focuses on the self-teacher's inherent capability to leverage this context to effectively guide diverse student rollouts. These two paradigms stabilize the performance gains of OPSD from complementary angles, suggesting a promising potential for integration.
\end{itemize}

\subsubsection{Relation with Methods integrating GRPO and SDPO}

Several concurrent studies have explored combining token-level OPSD with sequence-level GRPO~\cite{yang2026self, pan2026rlcsd}. In this section, we discuss both the similarities and differences between TESD and these approaches.

\begin{itemize}
    \item \textit{Similarity}. Both TESD and these methods integrate supervision at both the sequence level and token level.
    
    \item \textit{Difference}. \textit{(i)}~For the motivation, these concurrent studies degrade OPSD to an advantage-weighting component within GRPO. Conversely, TESD leverages GRPO to empower the self-teacher for improving OPSD's supervision quality. \textit{(ii)}~For the loss formulation, these studies adopt the standard GRPO objective, while TESD preserves the divergence-based loss of OPSD for the student model. \textit{(iii)}~For the optimization framework, these studies apply the GRPO and OPSD to a single student model. In contrast, TESD differentiates the model's dual roles, utilizing GRPO with specifically designed rewards for the self-teacher and OPSD with proposed adaptive weighting mechanism for the student, offering a principled framework rather than a naive objective blend.
\end{itemize}

\subsection{Ethical Considerations}
This work aims to improve the reasoning capabilities of large language models. We acknowledge that enhanced reasoning capabilities could also be misused for harmful purposes. Therefore, we emphasize the importance of deploying such models with robust safety alignment mechanisms to mitigate potential risks. All datasets used in our experiments are publicly available open-source resources. We adhere to their respective licenses and use them in accordance with their intended purposes.

\subsubsection{Limitations}

Despite outperforming representative token-level and sequence-level methods, TESD introduces additional hyperparameters, such as $\lambda_{\text{edit}}$, due to the inclusion of extra training objectives and rewards. 
Furthermore, TESD remains constrained by the teacher updating mechanism of the current OPSD paradigm~\cite{hubotter2026reinforcement}, which maintains a separate set of weights for the teacher and updates it via an exponential moving average to ensure training stability. This mechanism inherently smooths out the improvement of the teacher's capabilities, thereby limiting the speed at which the teacher evolves. 
In the future, to address these limitations, we will attempt to develop an adaptive independence term for the teacher reward and investigate the internal mechanism of the relationship between teacher updates and training stability.
Moreover, investigating whether TESD exhibits sparse update patterns~\cite{shen2024expanding, shen2026kernelized} offers an interesting direction for further understanding the student–teacher co-evolution framework.

%% file: Tables/rollout.tex
\begin{table}[]
\centering
\small
\begin{tabular}{@{}lccc@{}}
\toprule
Methods & $N=16$    & $N=32$    & $N=48$    \\ \midrule
SDPO    & 54.62 & 55.25 & 55.12 \\
TESD    & 54.87 & 56.63 & 59.62 \\ \bottomrule
\end{tabular}
\caption{Performance comparison of SDPO and TESD across different numbers of rollouts $N$. The experiments are conducted on the Biology subset of the SciKnowEval benchmark using Qwen3-4B over $10$ training epochs.}
\label{tab:rollout}
\end{table}

%% file: custom.bib
@misc{li2026rethinking,
    archiveprefix = {arXiv},
    author = {Li, Yaxuan and Zuo, Yuxin and He, Bingxiang and Zhang, Jinqian and Xiao, Chaojun and Qian, Cheng and Yu, Tianyu and Gao, Huan-ang and Yang, Wenkai and Liu, Zhiyuan and others},
    eprint = {2604.13016},
    title = {Rethinking On-Policy Distillation of Large Language Models: Phenomenology, Mechanism, and Recipe},
    url = {https://arxiv.org/abs/2604.13016},
    year = {2026}
}

@misc{hubotter2026reinforcement,
    archiveprefix = {arXiv},
    author = {H{\"u}botter, Jonas and L{\"u}beck, Frederike and Behric, Lejs and Baumann, Anton and Bagatella, Marco and Marta, Daniel and Hakimi, Ido and Shenfeld, Idan and Buening, Thomas Kleine and Guestrin, Carlos and others},
    eprint = {2601.20802},
    title = {Reinforcement Learning via Self-Distillation},
    url = {https://arxiv.org/abs/2601.20802},
    year = {2026}
}

@misc{yang2025qwen3,
    archiveprefix = {arXiv},
    author = {Yang, An and Li, Anfeng and Yang, Baosong and Zhang, Beichen and Hui, Binyuan and Zheng, Bo and Yu, Bowen and Gao, Chang and Huang, Chengen and Lv, Chenxu and others},
    eprint = {2505.09388},
    primaryclass = {cs.CL},
    title = {Qwen3 technical report},
    url = {https://arxiv.org/abs/2505.09388},
    year = {2025}
}

@misc{feng2024sciknoweval,
    archiveprefix = {arXiv},
    author = {Feng, Kehua and Shen, Xinyi and Wang, Weijie and Zhuang, Xiang and Tang, Yuqi and Zhang, Qiang and Ding, Keyan},
    eprint = {2406.09098},
    title = {Sciknoweval: Evaluating multi-level scientific knowledge of large language models},
    url = {https://arxiv.org/abs/2406.09098},
    year = {2024}
}

@misc{tang2023toolalpaca,
    archiveprefix = {arXiv},
    author = {Tang, Qiaoyu and Deng, Ziliang and Lin, Hongyu and Han, Xianpei and Liang, Qiao and Cao, Boxi and Sun, Le},
    eprint = {2306.05301},
    title = {Toolalpaca: Generalized tool learning for language models with 3000 simulated cases},
    url = {https://arxiv.org/abs/2306.05301},
    year = {2023}
}

@misc{zhao2026self,
    archiveprefix = {arXiv},
    author = {Zhao, Siyan and Xie, Zhihui and Liu, Mengchen and Huang, Jing and Pang, Guan and Chen, Feiyu and Grover, Aditya},
    eprint = {2601.18734},
    title = {Self-Distilled Reasoner: On-Policy Self-Distillation for Large Language Models},
    url = {https://arxiv.org/abs/2601.18734},
    year = {2026}
}

@misc{shenfeld2026self,
    archiveprefix = {arXiv},
    author = {Shenfeld, Idan and Damani, Mehul and H{\"u}botter, Jonas and Agrawal, Pulkit},
    eprint = {2601.19897},
    title = {Self-Distillation Enables Continual Learning},
    url = {https://arxiv.org/abs/2601.19897},
    year = {2026}
}

@misc{khatri2025art,
    archiveprefix = {arXiv},
    author = {Khatri, Devvrit and Madaan, Lovish and Tiwari, Rishabh and Bansal, Rachit and Duvvuri, Sai Surya and Zaheer, Manzil and Dhillon, Inderjit S and Brandfonbrener, David and Agarwal, Rishabh},
    eprint = {2510.13786},
    title = {The art of scaling reinforcement learning compute for llms},
    url = {https://arxiv.org/abs/2510.13786},
    year = {2025}
}

@inproceedings{yu2026dapo,
    author = {Qiying Yu and
Zheng Zhang and
Ruofei Zhu and
Yufeng Yuan and
Xiaochen Zuo and
Yu Yue and
Weinan Dai and
Tiantian Fan and
Gaohong Liu and
Juncai Liu and
Lingjun Liu and
Xin Liu and
Haibin Lin and
Zhiqi Lin and
Bole Ma and
Guangming Sheng and
Yuxuan Tong and
Chi Zhang and
Mofan Zhang and
Ru Zhang and
Wang Zhang and
Hang Zhu and
Jinhua Zhu and
Jiaze Chen and
Jiangjie Chen and
Chengyi Wang and
Hongli Yu and
Yuxuan Song and
Xiangpeng Wei and
Hao Zhou and
Jingjing Liu and
Wei{-}Ying Ma and
Ya{-}Qin Zhang and
Lin Yan and
Yonghui Wu and
Mingxuan Wang},
    bibsource = {dblp computer science bibliography, https://dblp.org},
    booktitle = {Advances in Neural Information Processing Systems 38: Annual Conference
on Neural Information Processing Systems 2025, NeurIPS 2025, San Diego,
CA, USA, December 2-7, 2025 / Mexico City, Mexico, November 30 - December
5, 2025},
    editor = {Danielle Belgrave and
Cheng Zhang and
Laura N. Montoya and
Hsuan{-}Tien Lin and
Razvan Pascanu and
Piotr Koniusz and
Marzyeh Ghassemi and
Nancy Chen and
Iv{\'{a}}n Vladimir Meza Ru{\'{\i}}z and
Arturo Loaiza{-}Bonilla},
    title = {{DAPO:} An Open-Source {LLM} Reinforcement Learning System at Scale},
    url = {http://papers.nips.cc/paper\_files/paper/2025/hash/a4277440d50f1f15d2cb4c14f7e0c0d2-Abstract-Conference.html},
    year = {2025}
}

@misc{liu2025understanding,
    archiveprefix = {arXiv},
    author = {Liu, Zichen and Chen, Changyu and Li, Wenjun and Qi, Penghui and Pang, Tianyu and Du, Chao and Lee, Wee Sun and Lin, Min},
    eprint = {2503.20783},
    title = {Understanding r1-zero-like training: A critical perspective},
    url = {https://arxiv.org/abs/2503.20783},
    year = {2025}
}

@article{yao2025your,
    author = {Yao, Feng and Liu, Liyuan and Zhang, Dinghuai and Dong, Chengyu and Shang, Jingbo and Gao, Jianfeng},
    journal = {URL https://fengyao. notion. site/off-policy-rl},
    title = {Your efficient rl framework secretly brings you off-policy rl training, August 2025},
    year = {2025}
}

@inproceedings{zhao2026reinforced,
    address = {San Diego, California, United States},
    author = {Zhao, Ziqi  and
Ren, Zhaochun  and
Zou, Jiahong  and
Yang, Liu  and
Xu, Zhiwei  and
Ge, Xuri  and
Chen, Zhumin  and
Ma, Xinyu  and
Shi, Daiting  and
Wang, Shuaiqiang  and
Yin, Dawei  and
Xin, Xin},
    booktitle = {Proceedings of the 64th Annual Meeting of the {A}ssociation for {C}omputational {L}inguistics (Volume 1: Long Papers)},
    doi = {10.18653/v1/2026.acl-long.2216},
    editor = {Liakata, Maria  and
Moreira, Viviane P.  and
Zhang, Jiajun  and
Jurgens, David},
    isbn = {979-8-89176-390-6},
    pages = {47994--48007},
    publisher = {Association for Computational Linguistics},
    title = {Reinforced Efficient Reasoning via Semantically Diverse Exploration},
    url = {https://aclanthology.org/2026.acl-long.2216/},
    year = {2026}
}

@misc{guo2025deepseek,
    archiveprefix = {arXiv},
    author = {Guo, Daya and Yang, Dejian and Zhang, Haowei and Song, Junxiao and Zhang, Ruoyu and Xu, Runxin and Zhu, Qihao and Ma, Shirong and Wang, Peiyi and Bi, Xiao and others},
    eprint = {2501.12948},
    title = {Deepseek-r1: Incentivizing reasoning capability in llms via reinforcement learning},
    url = {https://arxiv.org/abs/2501.12948},
    year = {2025}
}

@misc{shao2024deepseekmath,
    archiveprefix = {arXiv},
    author = {Shao, Zhihong and Wang, Peiyi and Zhu, Qihao and Xu, Runxin and Song, Junxiao and Bi, Xiao and Zhang, Haowei and Zhang, Mingchuan and Li, YK and Wu, Yang and others},
    eprint = {2402.03300},
    primaryclass = {cs.CL},
    title = {Deepseekmath: Pushing the limits of mathematical reasoning in open language models},
    url = {https://arxiv.org/abs/2402.03300},
    year = {2024}
}

@inproceedings{yu2025dapo,
    author = {Qiying Yu and
Zheng Zhang and
Ruofei Zhu and
Yufeng Yuan and
Xiaochen Zuo and
Yu Yue and
Weinan Dai and
Tiantian Fan and
Gaohong Liu and
Juncai Liu and
Lingjun Liu and
Xin Liu and
Haibin Lin and
Zhiqi Lin and
Bole Ma and
Guangming Sheng and
Yuxuan Tong and
Chi Zhang and
Mofan Zhang and
Ru Zhang and
Wang Zhang and
Hang Zhu and
Jinhua Zhu and
Jiaze Chen and
Jiangjie Chen and
Chengyi Wang and
Hongli Yu and
Yuxuan Song and
Xiangpeng Wei and
Hao Zhou and
Jingjing Liu and
Wei{-}Ying Ma and
Ya{-}Qin Zhang and
Lin Yan and
Yonghui Wu and
Mingxuan Wang},
    bibsource = {dblp computer science bibliography, https://dblp.org},
    booktitle = {Advances in Neural Information Processing Systems 38: Annual Conference
on Neural Information Processing Systems 2025, NeurIPS 2025, San Diego,
CA, USA, December 2-7, 2025 / Mexico City, Mexico, November 30 - December
5, 2025},
    editor = {Danielle Belgrave and
Cheng Zhang and
Laura N. Montoya and
Hsuan{-}Tien Lin and
Razvan Pascanu and
Piotr Koniusz and
Marzyeh Ghassemi and
Nancy Chen and
Iv{\'{a}}n Vladimir Meza Ru{\'{\i}}z and
Arturo Loaiza{-}Bonilla},
    title = {{DAPO:} An Open-Source {LLM} Reinforcement Learning System at Scale},
    url = {http://papers.nips.cc/paper\_files/paper/2025/hash/a4277440d50f1f15d2cb4c14f7e0c0d2-Abstract-Conference.html},
    year = {2025}
}

@misc{zheng2025group,
    archiveprefix = {arXiv},
    author = {Zheng, Chujie and Liu, Shixuan and Li, Mingze and Chen, Xiong-Hui and Yu, Bowen and Gao, Chang and Dang, Kai and Liu, Yuqiong and Men, Rui and Yang, An and others},
    eprint = {2507.18071},
    title = {Group sequence policy optimization},
    url = {https://arxiv.org/abs/2507.18071},
    year = {2025}
}

@misc{chu2025gpg,
    archiveprefix = {arXiv},
    author = {Chu, Xiangxiang and Huang, Hailang and Zhang, Xiao and Wei, Fei and Wang, Yong},
    eprint = {2504.02546},
    title = {Gpg: A simple and strong reinforcement learning baseline for model reasoning},
    url = {https://arxiv.org/abs/2504.02546},
    year = {2025}
}

@inproceedings{gu2024minillm,
    author = {Yuxian Gu and
Li Dong and
Furu Wei and
Minlie Huang},
    bibsource = {dblp computer science bibliography, https://dblp.org},
    booktitle = {The Twelfth International Conference on Learning Representations,
{ICLR} 2024, Vienna, Austria, May 7-11, 2024},
    publisher = {OpenReview.net},
    title = {MiniLLM: Knowledge Distillation of Large Language Models},
    url = {https://openreview.net/forum?id=5h0qf7IBZZ},
    year = {2024}
}

@misc{kumar2025llm,
    archiveprefix = {arXiv},
    author = {Kumar, Komal and Ashraf, Tajamul and Thawakar, Omkar and Anwer, Rao Muhammad and Cholakkal, Hisham and Shah, Mubarak and Yang, Ming-Hsuan and Torr, Phillip HS and Khan, Fahad Shahbaz and Khan, Salman},
    eprint = {2502.21321},
    title = {Llm post-training: A deep dive into reasoning large language models},
    url = {https://arxiv.org/abs/2502.21321},
    year = {2025}
}

@misc{achiam2023gpt,
    archiveprefix = {arXiv},
    author = {Achiam, Josh and Adler, Steven and Agarwal, Sandhini and Ahmad, Lama and Akkaya, Ilge and Aleman, Florencia Leoni and Almeida, Diogo and Altenschmidt, Janko and Altman, Sam and Anadkat, Shyamal and others},
    eprint = {2303.08774},
    title = {Gpt-4 technical report},
    url = {https://arxiv.org/abs/2303.08774},
    year = {2023}
}

@inproceedings{agarwal2024policy,
    author = {Rishabh Agarwal and
Nino Vieillard and
Yongchao Zhou and
Piotr Stanczyk and
Sabela Ramos Garea and
Matthieu Geist and
Olivier Bachem},
    bibsource = {dblp computer science bibliography, https://dblp.org},
    booktitle = {The Twelfth International Conference on Learning Representations,
{ICLR} 2024, Vienna, Austria, May 7-11, 2024},
    publisher = {OpenReview.net},
    title = {On-Policy Distillation of Language Models: Learning from Self-Generated
Mistakes},
    url = {https://openreview.net/forum?id=3zKtaqxLhW},
    year = {2024}
}

@article{lu2025onpolicydistillation,
    author = {Kevin Lu and Thinking Machines Lab},
    doi = {10.64434/tml.20251026},
    journal = {Thinking Machines Lab: Connectionism},
    note = {https://thinkingmachines.ai/blog/on-policy-distillation},
    title = {On-Policy Distillation},
    year = {2025}
}

@misc{yang2026self,
    archiveprefix = {arXiv},
    author = {Yang, Chenxu and Qin, Chuanyu and Si, Qingyi and Chen, Minghui and Gu, Naibin and Yao, Dingyu and Lin, Zheng and Wang, Weiping and Wang, Jiaqi and Duan, Nan},
    eprint = {2604.03128},
    title = {Self-distilled rlvr},
    url = {https://arxiv.org/abs/2604.03128},
    year = {2026}
}

@misc{yu2026multi,
    archiveprefix = {arXiv},
    author = {Yu, Weichen and Li, Xiaomin and Zhao, Yizhou and Liu, Xiaoze and Zhang, Ruowang and Wang, Haixin and Luo, Yinyi and Wu, Chen Henry and Mittal, Gaurav and Fredrikson, Matt and others},
    eprint = {2605.12652},
    title = {Multi-Rollout On-Policy Distillation via Peer Successes and Failures},
    url = {https://arxiv.org/abs/2605.12652},
    year = {2026}
}

@misc{olmo2025olmo,
    archiveprefix = {arXiv},
    author = {Olmo, Team and Ettinger, Allyson and Bertsch, Amanda and Kuehl, Bailey and Graham, David and Heineman, David and Groeneveld, Dirk and Brahman, Faeze and Timbers, Finbarr and Ivison, Hamish and others},
    eprint = {2512.13961},
    primaryclass = {cs.CL},
    title = {Olmo 3},
    url = {https://arxiv.org/abs/2512.13961},
    year = {2025}
}

@inproceedings{jain2025livecodebench,
    author = {Naman Jain and
King Han and
Alex Gu and
Wen{-}Ding Li and
Fanjia Yan and
Tianjun Zhang and
Sida Wang and
Armando Solar{-}Lezama and
Koushik Sen and
Ion Stoica},
    bibsource = {dblp computer science bibliography, https://dblp.org},
    booktitle = {The Thirteenth International Conference on Learning Representations,
{ICLR} 2025, Singapore, April 24-28, 2025},
    publisher = {OpenReview.net},
    title = {LiveCodeBench: Holistic and Contamination Free Evaluation of Large
Language Models for Code},
    url = {https://openreview.net/forum?id=chfJJYC3iL},
    year = {2025}
}

@misc{zeng2026glm,
    archiveprefix = {arXiv},
    author = {Zeng, Aohan and Lv, Xin and Hou, Zhenyu and Du, Zhengxiao and Zheng, Qinkai and Chen, Bin and Yin, Da and Ge, Chendi and Huang, Chenghua and Xie, Chengxing and others},
    eprint = {2602.15763},
    title = {Glm-5: from vibe coding to agentic engineering},
    url = {https://arxiv.org/abs/2602.15763},
    year = {2026}
}

@misc{jin2026entropy,
    archiveprefix = {arXiv},
    author = {Jin, Woogyeol and Min, Taywon and Yang, Yongjin and Kadhe, Swanand Ravindra and Zhou, Yi and Wei, Dennis and Baracaldo, Nathalie and Lee, Kimin},
    eprint = {2603.07079},
    primaryclass = {cs.LG},
    title = {Entropy-Aware On-Policy Distillation of Language Models},
    url = {https://arxiv.org/abs/2603.07079},
    year = {2026}
}

@misc{yuan2023scaling,
    archiveprefix = {arXiv},
    author = {Yuan, Zheng and Yuan, Hongyi and Li, Chengpeng and Dong, Guanting and Lu, Keming and Tan, Chuanqi and Zhou, Chang and Zhou, Jingren},
    eprint = {2308.01825},
    title = {Scaling relationship on learning mathematical reasoning with large language models},
    url = {https://arxiv.org/abs/2308.01825},
    year = {2023}
}

@misc{Zhou2023InstructionFollowingEF,
    archiveprefix = {arXiv},
    author = {Jeffrey Zhou and Tianjian Lu and Swaroop Mishra and Siddhartha Brahma and Sujoy Basu and Yi Luan and Denny Zhou and Le Hou},
    eprint = {2311.07911},
    primaryclass = {cs.CL},
    title = {Instruction-Following Evaluation for Large Language Models},
    url = {https://arxiv.org/abs/2311.07911},
    year = {2023}
}

@inproceedings{Li2023HaluEvalAL,
    address = {Singapore},
    author = {Li, Junyi  and
Cheng, Xiaoxue  and
Zhao, Xin  and
Nie, Jian-Yun  and
Wen, Ji-Rong},
    booktitle = {Proceedings of the 2023 Conference on Empirical Methods in Natural Language Processing},
    doi = {10.18653/v1/2023.emnlp-main.397},
    editor = {Bouamor, Houda  and
Pino, Juan  and
Bali, Kalika},
    pages = {6449--6464},
    publisher = {Association for Computational Linguistics},
    title = {{H}alu{E}val: A Large-Scale Hallucination Evaluation Benchmark for Large Language Models},
    url = {https://aclanthology.org/2023.emnlp-main.397/},
    year = {2023}
}

@inproceedings{Lin2021TruthfulQAMH,
    address = {Dublin, Ireland},
    author = {Lin, Stephanie  and
Hilton, Jacob  and
Evans, Owain},
    booktitle = {Proceedings of the 60th Annual Meeting of the Association for Computational Linguistics (Volume 1: Long Papers)},
    doi = {10.18653/v1/2022.acl-long.229},
    editor = {Muresan, Smaranda  and
Nakov, Preslav  and
Villavicencio, Aline},
    pages = {3214--3252},
    publisher = {Association for Computational Linguistics},
    title = {{T}ruthful{QA}: Measuring How Models Mimic Human Falsehoods},
    url = {https://aclanthology.org/2022.acl-long.229/},
    year = {2022}
}

@inproceedings{Verschueren2018Preprint,
    author = {Sarah Verschueren and J. van Aalst and A. Bangels and J. Toelen and K. Allegaert and C. Buffel and G. Vander Stichele},
    title = {(Preprint)},
    year = {2018}
}

@article{Hendrycks2021MeasuringMP,
    author = {Dan Hendrycks and Collin Burns and Saurav Kadavath and Akul Arora and Steven Basart and Eric Tang and D. Song and J. Steinhardt},
    booktitle = {NeurIPS Datasets and Benchmarks},
    journal = {ArXiv},
    title = {Measuring Mathematical Problem Solving With the MATH Dataset},
    volume = {abs/2103.03874},
    year = {2021}
}

@misc{Clark2018ThinkYH,
    archiveprefix = {arXiv},
    author = {Peter Clark and Isaac Cowhey and Oren Etzioni and Tushar Khot and Ashish Sabharwal and Carissa Schoenick and Oyvind Tafjord},
    eprint = {1803.05457},
    primaryclass = {cs.AI},
    title = {Think you have Solved Question Answering? Try ARC, the AI2 Reasoning Challenge},
    url = {https://arxiv.org/abs/1803.05457},
    year = {2018}
}

@misc{Cobbe2021TrainingVT,
    archiveprefix = {arXiv},
    author = {Karl Cobbe and Vineet Kosaraju and Mo Bavarian and Mark Chen and Heewoo Jun and Lukasz Kaiser and Matthias Plappert and Jerry Tworek and Jacob Hilton and Reiichiro Nakano and Christopher Hesse and John Schulman},
    eprint = {2110.14168},
    title = {Training Verifiers to Solve Math Word Problems},
    url = {https://arxiv.org/abs/2110.14168},
    year = {2021}
}

@misc{Chen2021EvaluatingLL,
    archiveprefix = {arXiv},
    author = {Mark Chen and Jerry Tworek and Heewoo Jun and Qiming Yuan and Henrique Pond{\'e} and Jared Kaplan and Harrison Edwards and Yura Burda and Nicholas Joseph and Greg Brockman and Alex Ray and Raul Puri and Gretchen Krueger and Michael Petrov and Heidy Khlaaf and Girish Sastry and Pamela Mishkin and Brooke Chan and Scott Gray and Nick Ryder and Mikhail Pavlov and Alethea Power and Lukasz Kaiser and Mo Bavarian and Clemens Winter and Phil Tillet and Felipe Petroski Such and David W. Cummings and Matthias Plappert and Fotios Chantzis and Elizabeth Barnes and Ariel Herbert-Voss and William H. Guss and Alex Nichol and Igor Babuschkin and Suchir Balaji and Shantanu Jain and Andrew Carr and Jan Leike and Josh Achiam and Vedant Misra and Evan Morikawa and Alec Radford and Matthew M. Knight and Miles Brundage and Mira Murati and Katie Mayer and Peter Welinder and Bob McGrew and Dario Amodei and Sam McCandlish and Ilya Sutskever and Wojciech Zaremba},
    eprint = {2107.03374},
    primaryclass = {cs.LG},
    title = {Evaluating Large Language Models Trained on Code},
    url = {https://arxiv.org/abs/2107.03374},
    year = {2021}
}

@misc{sheng2024hybridflow,
    archiveprefix = {arXiv},
    author = {Guangming Sheng and Chi Zhang and Zilingfeng Ye and Xibin Wu and Wang Zhang and Ru Zhang and Yanghua Peng and Haibin Lin and Chuan Wu},
    eprint = {2409.19256},
    primaryclass = {cs.LG},
    title = {HybridFlow: A Flexible and Efficient RLHF Framework},
    url = {https://arxiv.org/abs/2409.19256},
    year = {2024}
}

@misc{eval-harness,
    author = {Gao, Leo and Tow, Jonathan and Abbasi, Baber and Biderman, Stella and Black, Sid and DiPofi, Anthony and Foster, Charles and Golding, Laurence and Hsu, Jeffrey and Le Noac'h, Alain and Li, Haonan and McDonell, Kyle and Muennighoff, Niklas and Ociepa, Chris and Phang, Jason and Reynolds, Laria and Schoelkopf, Hailey and Skowron, Aviya and Sutawika, Lintang and Tang, Eric and Thite, Anish and Wang, Ben and Wang, Kevin and Zou, Andy},
    doi = {10.5281/zenodo.12608602},
    publisher = {Zenodo},
    title = {The Language Model Evaluation Harness},
    url = {https://zenodo.org/records/12608602},
    version = {v0.4.3},
    year = {2024}
}

@misc{pan2026rlcsd,
    archiveprefix = {arXiv},
    author = {Pan, Leyi and Tao, Shuchang and Zhai, Yunpeng and Zhang, Lingzhe and Liu, Zhaoyang and Ding, Bolin and Liu, Aiwei and Wen, Lijie},
    eprint = {2606.11709},
    title = {RLCSD: Reinforcement Learning with Contrastive On-Policy Self-Distillation},
    url = {https://arxiv.org/abs/2606.11709},
    year = {2026}
}

@inproceedings{zheng2023judging,
    author = {Lianmin Zheng and
Wei{-}Lin Chiang and
Ying Sheng and
Siyuan Zhuang and
Zhanghao Wu and
Yonghao Zhuang and
Zi Lin and
Zhuohan Li and
Dacheng Li and
Eric P. Xing and
Hao Zhang and
Joseph E. Gonzalez and
Ion Stoica},
    bibsource = {dblp computer science bibliography, https://dblp.org},
    booktitle = {Advances in Neural Information Processing Systems 36: Annual Conference
on Neural Information Processing Systems 2023, NeurIPS 2023, New Orleans,
LA, USA, December 10 - 16, 2023},
    editor = {Alice Oh and
Tristan Naumann and
Amir Globerson and
Kate Saenko and
Moritz Hardt and
Sergey Levine},
    title = {Judging LLM-as-a-Judge with MT-Bench and Chatbot Arena},
    url = {http://papers.nips.cc/paper\_files/paper/2023/hash/91f18a1287b398d378ef22505bf41832-Abstract-Datasets\_and\_Benchmarks.html},
    year = {2023}
}

@misc{shen2026policy,
    archiveprefix = {arXiv},
    author = {Shen, Shufan and Hou, Zhongni and Sun, Junshu and Zhang, Yufei and Lin, Wei and Yin, Guojun and Huang, Qingming and Wang, Shuhui},
    eprint = {2609.36659},
    title = {On-Policy Parameter Update Direction Underlies Generalization in LLM Post-Training},
    url = {https://arxiv.org/abs/2609.36659},
    year = {2026}
}

@article{shen2026vl,
  title={Vl-sae: Interpreting and enhancing vision-language alignment with a unified concept set},
  author={Shen, Shufan and Sun, Junshu and Huang, Qingming and Wang, Shuhui},
  journal={Advances in Neural Information Processing Systems},
  volume={38},
  pages={45235--45265},
  year={2026}
}

@article{shen2026kernelized,
  title={Kernelized sparse fine-tuning with bi-level parameter competition for vision models},
  author={Shen, Shufan and Sun, Junshu and Wang, Shuhui and Huang, Qingming},
  journal={IEEE Transactions on Pattern Analysis and Machine Intelligence},
  year={2026},
  publisher={IEEE}
}

@article{shen2024expanding,
  title={Expanding sparse tuning for low memory usage},
  author={Shen, Shufan and Sun, Junshu and Ji, Xiangyang and Huang, Qingming and Wang, Shuhui},
  journal={Advances in Neural Information Processing Systems},
  volume={37},
  pages={76616--76642},
  year={2024}
}
